\documentclass{article}

\usepackage{microtype}
\usepackage{graphicx}
\usepackage{subcaption}
\usepackage{booktabs} 
\usepackage{multirow}

\usepackage{hyperref}

\usepackage[preprint]{icml2026}

\usepackage{amsmath}
\usepackage{amssymb}
\usepackage{mathtools}
\usepackage{amsthm}
\usepackage{algorithm}
\usepackage{algorithmic}

\usepackage[capitalize,noabbrev]{cleveref}
\usepackage{wrapfig}

\theoremstyle{plain}
\newtheorem{theorem}{Theorem}[section]
\newtheorem{proposition}[theorem]{Proposition}

\newtheorem{corollary}[theorem]{Corollary}
\theoremstyle{definition}

\newtheorem{assumption}[theorem]{Assumption}
\theoremstyle{remark}
\newtheorem{remark}[theorem]{Remark}
\crefname{assumption}{Assumption}{Assumptions}
\crefname{remark}{Remark}{Remarks}

\DeclareMathSizes{10}{9}{6.5}{5}

\AtBeginDocument{\ifdim\parskip>2pt\relax\addtolength{\parskip}{-2pt}\else\setlength{\parskip}{0pt}\fi}

\icmltitlerunning{An Augmented Lagrangian Approach to Lifted Bellman Linear Programming}

\begin{document}

\twocolumn[
  \icmltitle{An Augmented Lagrangian Approach to Lifted Bellman Linear Programming}

  \icmlsetsymbol{equal}{*}

  \begin{icmlauthorlist}
    \icmlauthor{Jongchan Park}{comp}
    \icmlauthor{Hyukjun Yang}{equal,sch}
    \icmlauthor{Narim Jeong}{equal,sch}
    \icmlauthor{Donghwan Lee}{sch}
  \end{icmlauthorlist}

  \icmlaffiliation{comp}{NAVER LABS, Seongnam, Republic of Korea}
  \icmlaffiliation{sch}{Department of EE, Korea Advanced Institute of Science and Technology (KAIST), Daejeon, Republic of Korea}

  \icmlcorrespondingauthor{Donghwan Lee}{donghwan@kaist.ac.kr}

  \icmlkeywords{Machine Learning, ICML}

  \vskip 0.3in
]

\printAffiliationsAndNotice{\icmlEqualContribution}

\begin{abstract}
Temporal-difference critics regress onto a bootstrapped target and usually need a target network. Lifted Bellman Linear Programming (LBLP) instead expresses the in-sample optimal values of offline RL as the solution to a linear program over the dataset. Its neural solver ALBUM relaxes the constraints into hinge penalties with fixed coefficients, and its guarantees are conditions on coefficients chosen in advance. We solve the same program with the method of multipliers: the coefficients become multipliers learned by dual ascent, with one penalty parameter $\rho$. In the convex case the fixed points are the KKT pairs of LBLP, and at every stationary point the applied constraint weights balance the objective exactly. The critic can serve a Gaussian actor, a flow actor, and best-of-$N$ sampling, and for the Gaussian actor we propose an advantage-normalized DDPG+BC objective. With a single $Q$ and $V$ and no target network, our AL-LBLP with best-of-$N$ reaches 60.1\% average success on the 50 state-based OGBench tasks. On long-horizon tasks, it reaches 81.5\% on \textit{humanoidmaze-giant} and 79.8\% on \textit{antmaze-giant}. It does not collapse up to batch size 8192.
\end{abstract}
\section{Introduction}
\label{sec:intro}

Offline RL learns a policy from a fixed dataset without further interaction \citep{levine2020offline}. The critic is usually trained by temporal-difference (TD) regression onto a bootstrapped target. That target evaluates $Q$ at next actions from the learning policy, and the Bellman maximization accumulates errors at the actions the data do not cover \citep{fujimoto2019off,kumar2020conservative}. Therefore prior work constrains the policy or penalizes such actions, and stabilizes the regression with target networks and critic ensembles.

Linear-programming (LP) formulations of Bellman optimality have been developed to address these limitations. TD learning with function approximation can diverge off-policy \citep{baird1995residual,tsitsiklis1997analysis}. Furthermore, Q-learning with general function approximation lacks convergence guarantees, whereas LP-based convex relaxations provide such guarantees \citep{mehta2020convex,lee2020unified}. However, on deep offline RL benchmarks, LP approaches had not competed with TD-based methods. Dual density-ratio methods were evaluated on D4RL with mixed results across tasks \citep{lee2021optidice,sikchi2024dual}. To our knowledge, ALBUM \citep{yang2026lifted} and our AL-LBLP are the first primal LP methods whose performance on such a benchmark matches that of recent TD algorithms.

Both methods build on Lifted Bellman Linear Programming (LBLP), introduced by \citet{yang2026lifted}. It characterizes the in-sample Bellman-optimal pair $(Q^*_\mathcal{D},V^*_\mathcal{D})$ as the solution of a linear program whose constraints are defined over dataset transitions. Therefore nothing is regressed onto a moving target, and no target network is needed. Furthermore, this formulation helps to avoid the compounding error associated with repeated bootstrapped regression \citep{lee2021optidice,yang2026lifted}. Its neural solver ALBUM minimizes a hinge relaxation of this program with fixed coefficients and matches the OGBench average of FQL with a single critic. The guarantees of ALBUM, however, are conditions on coefficients chosen in advance. In particular, exactness is guaranteed only above dual thresholds that depend on the problem and are not known beforehand. AL-LBLP, the method of this paper, learns the multipliers and removes this choice.

A critic that does not require an actor during training also matters beyond the benchmark. Recent methods that steer or fine-tune generalist robot policies train their critics by TD learning, stabilized with target networks or ensembles \citep{nakamoto2024steering,dong2025expo,singh2026seeq}. SeeQ learns its vision-language $Q$ function with a target network, and EXPO \citep{dong2025expo} uses an ensemble of ten $Q$ networks. A TD target $Q(s',\pi(s'))$ calls the policy at every critic update. With a vision-language-action (VLA) model as the policy, every critic step then pays the policy's forward pass and depends on actions outside the data. The LBLP critic bootstraps through $V(s')$ at dataset transitions, so it is trained without the policy and can be attached to any policy afterwards. For the same reason, several actors, of different types or with different values of $\alpha$, can share one critic and be trained together in one run. For example, six values of $\alpha$ can be trained together, 4.3$\times$ faster than in separate runs (\cref{app:impl}). Our experiments on OGBench demonstrate that one critic serves three policy extractions well (\cref{tab:main}), and training does not collapse when the batch size grows to 8192 (\cref{tab:longhorizon}).

We solve LBLP with the method of multipliers, the augmented Lagrangian method for constrained programs \citep{hestenes1969multiplier,powell1969method,bertsekas1982constrained}. Each constraint family carries a multiplier that is learned by dual ascent. One penalty parameter $\rho$ sets both the dual step size and the curvature of the penalty on violated rows. The weight that a row receives, $[\mu+\rho y]_{+}$, grows with its violation $y$, whereas a hinge gives every violated row the same weight. For the actor we propose a DDPG+BC objective whose $Q$ term is normalized by the mean absolute advantage $|Q-V|$ rather than by $|Q|$. Our contributions are:
\begin{itemize}
  \item \textbf{Same program, different solver.} We propose AL-LBLP, which replaces the three fixed hinge coefficients of ALBUM by learned multipliers with one penalty parameter $\rho$ (\cref{sec:alm}). Like $\alpha$, $\rho$ is selected per environment (\cref{sec:experiments}). In the convex case its fixed points are the KKT pairs of LBLP. At every stationary point the applied row weights satisfy a mass identity that turns the bounded-update condition of ALBUM into an equality (\cref{sec:theory}).
  \item \textbf{Advantage-normalized actor.} We divide the $Q$ term of DDPG+BC by $\operatorname{sg}\,\mathbb{E}|Q(s,\pi(s))-V(s)|$, which the LP critics provide directly (\cref{sec:actor}). The update is invariant to positive affine maps of the reward and to common shifts of $Q$ and $V$, unlike the $|Q|$ normalization of TD3+BC. On the same critic it raises the best single-$\alpha$ mean from 29.5 to 49.3 and narrows the range of useful $\alpha$ (\cref{sec:norm}).
  \item \textbf{Results.} With a single $Q$, $V$ and no target network or ensemble, the same critic reaches 55.8 with a Gaussian actor, 55.7 with a flow actor, and 60.1 with the best-of-$N$ policy on the 50 state-based OGBench tasks. These are the highest averages of the Gaussian and best-of-$N$ groups and the second of the flow group (\cref{sec:main}). With $K=50$ and $\gamma=0.999$ it reaches 81.5\% on \textit{humanoidmaze-giant} and 79.8\% on \textit{antmaze-giant} (8 seeds, \cref{sec:longhorizon}). It does not collapse at batch sizes up to 8192, whereas TD-based FQL and FAC fall to 0.0 at batch 2048--4096 (\cref{tab:batch-fql}).
  \item \textbf{Analysis.} $\rho$ trades early training speed against long-run stability (\cref{sec:rho}). Against ALBUM, the gap lies in the $K$-step constraints at the goal, which accelerate value propagation: on the OGBench giant mazes their mean violation is 471 for ALBUM but only 0.9 for AL-LBLP (\cref{sec:album-cmp}).
\end{itemize}

\section{Related Work}
\label{sec:related}

\paragraph{Offline RL.}
Offline RL learns a policy from a fixed dataset without further interaction \citep{levine2020offline}. Its main difficulty is that the Bellman maximization amplifies value errors at actions the data do not contain \citep{fujimoto2019off,kumar2020conservative}. To address this, conservative critics penalize the values of out-of-distribution actions \citep{kumar2020conservative}, and actor regularization keeps the policy close to the behavior policy \citep{fujimoto2019off,fujimoto2021minimalist,tarasov2023revisiting}. In-sample methods such as IQL instead learn the critic only from dataset actions \citep{kostrikov2021offline}. More recently, diffusion and flow policies model the behavior distribution expressively \citep{hansenestruch2023idql,park2025flow,chae2026flow,mu2026preserve}. They keep the policy near the data implicitly, by sampling from a behavior-cloning flow or by distilling the policy from it. These methods learn the critic by TD regression onto a bootstrapped target, usually with target networks and critic ensembles. In contrast, our AL-LBLP takes an LP approach: its critic is learned in-sample, without a target network or an ensemble. Nevertheless, it performs on par with or better than recent TD-based methods that rely on both (\cref{sec:main}).

\paragraph{Linear programs for Bellman optimality.}
Writing Bellman optimality as a linear program removes the regression onto a bootstrapped target \citep{defarias2003linear,puterman2014markov}. This made LP approaches a way around the weaknesses of TD learning. With function approximation, TD learning can diverge off-policy \citep{baird1995residual,tsitsiklis1997analysis}, and its convergence is known only under restrictive conditions \citep{mehta2020convex,jeong2018primal,lee2020unified}. Approximate LP and its primal--dual solvers develop this view with theoretical guarantees \citep{neu2023efficient,ozdaglar2023revisiting}. In offline RL, most LP methods pass to the dual, estimate stationary density ratios, and recover the policy from them \citep{nachum2020reinforcement,lee2021optidice,sikchi2024dual}. For example, OptiDICE has no compounding error from bootstrapped estimates and stays stable as $\gamma$ grows, where CQL does not \citep{lee2021optidice}. On D4RL, however, these methods matched TD-based methods only on some tasks. Logistic Q-learning also starts from a regularized dual and recovers a primal loss by conjugacy \citep{basserrano2020logistic}. Our augmented Lagrangian has the same form, with a quadratic proximal term on the multipliers, but it learns the multipliers explicitly and keeps $Q$ and $V$ as the primal variables.

\paragraph{Lifted Bellman Linear Programming.}
\citet{yang2026lifted} proposed LBLP, which lifts the LP to the pair $(Q,V)$ so that every constraint involves only dataset pairs. Because $Q$ and $V$ are its variables, the policy is extracted directly from the learned $Q$ and $V$, as in actor--critic methods. They solved LBLP with neural networks by relaxing its constraints into hinge penalties with fixed coefficients $(\lambda_{\mathrm{B}},\lambda_{\mathrm{K}},\lambda_{\mathrm{E}})$. The resulting method, ALBUM, matches the OGBench average of FQL with a single critic and no target network. Its guarantees are conditions on these coefficients: exactness holds above problem-dependent dual thresholds, and a bounded detached update needs a separate inequality. We build AL-LBLP on LBLP and change how the program is solved. The fixed coefficients become multipliers learned by dual ascent, with a single penalty parameter $\rho$. This is the method of multipliers, the classical solver for constrained programs \citep{hestenes1969multiplier,powell1969method,rockafellar1973dual,bertsekas1982constrained}, which, to our knowledge, is new for a Bellman LP with neural $Q$ and $V$. Like ALBUM, AL-LBLP is in-sample and uses a single $Q$ and $V$ without a target network. To our knowledge, ALBUM and AL-LBLP are the first primal LP methods to reach the level of recent TD-based methods on deep offline benchmarks. With a Gaussian actor, AL-LBLP improves on ALBUM by about 29\%, and the gap is largest on long-horizon tasks (\cref{sec:longhorizon}).

\section{Preliminaries}
\label{sec:prelim}

\paragraph{Markov decision processes and offline RL.}
We consider a discounted MDP $\mathcal{M}=(\mathcal{S},\mathcal{A},P,R,\gamma)$ with transition kernel $P(\cdot\mid s,a)$, expected reward $R(s,a)$, and discount $0\le\gamma<1$. The optimal value functions satisfy $V^*(s)=\max_a Q^*(s,a)$. In the offline setting the agent is given a fixed dataset $\mathcal{D}=\{(s_i,a_i,r_i,s_i')\}_{i=1}^n$ collected by a behavior policy $\beta$. The data are contiguous trajectories. We write $\mathcal{M}_\mathcal{D}$ for the empirical MDP they induce, whose actions at $s$ are those the data contain at $s$. As in \citet[App.~C]{yang2026lifted}, a successor state that the data never leave is treated as absorbing with value zero. For a window of $K$ consecutive transitions starting at $(s,a)=(s_0,a_0)$, we write $\hat y_K(s,a)=\sum_{t<K}\gamma^t r_t+\gamma^K V(s_K)$ for the $K$-step return bootstrapped at its end. Terminal masks are absorbed into $\gamma$ and $\gamma^K$. $\operatorname{sg}[\cdot]$ denotes \texttt{stop\_gradient}, $[x]_+=\max\{x,0\}$, and $\mathbb{E}_\mathcal{D}$ or $\mathbb{E}_i$ the empirical expectation over the dataset.

\paragraph{Linear programming characterization of Bellman optimality.}
For any weight $c$ with strictly positive entries, $V^*$ is the unique solution of $\min_V\langle c,V\rangle$ subject to $V(s)\ge R(s,a)+\gamma(PV)(s,a)$ for all $(s,a)$ \citep{defarias2003linear,puterman2014markov}. Neither this program nor its action-value counterpart carries over to offline RL directly. The state-value form determines only $V^*$, so a policy must be recovered from the dual. The action-value form imposes a constraint for every action at the next state, including actions the data do not contain.

\paragraph{Lifted Bellman Linear Program.}
LBLP \citep{yang2026lifted} resolves both difficulties by lifting the program to the pair $(Q,V)$ through the epigraph inequality $Q(s,a)\le V(s)$. On $\mathcal{M}_\mathcal{D}$, with weights $\omega_{Q},\omega_{V}>0$ and rollout length $K$, the program is
\begin{equation}
  \label{eq:lblp}
  \begin{aligned}
  \min_{Q,V}\;&\mathbb{E}_\mathcal{D}[\omega_{Q} Q(s,a)+\omega_{V} V(s)]\\
  \text{s.t.}\;&g_{\mathrm{B}}\le0,\quad g_{\mathrm{E}}\le0,\quad g_{\mathrm{KQ}}\le0,\quad g_{\mathrm{KV}}\le0,
  \end{aligned}
\end{equation}
with constraint functions, each indexed by a dataset pair,
\begin{align*}
  g_{\mathrm{B}}(s,a)&=r+\gamma V(s')-Q(s,a),\\
  g_{\mathrm{E}}(s,a)&=Q(s,a)-V(s),\\
  g_{\mathrm{KQ}}(s,a)&=\hat y_K(s,a)-Q(s,a),\\
  g_{\mathrm{KV}}(s,a)&=\hat y_K(s,a)-V(s).
\end{align*}
Its unique minimizer is the in-sample Bellman-optimal pair $(Q^*_\mathcal{D},V^*_\mathcal{D})$ of $\mathcal{M}_\mathcal{D}$. The rollout constraints $g_{\mathrm{KQ}},g_{\mathrm{KV}}$ are implied by the one-step constraints for the sampled $\hat y_K$ under deterministic dynamics, so they leave the solution unchanged for any $K$. In an idealized coordinate-wise iteration they accelerate value propagation \citep[Lemma~4.2, Prop.~4.3, Thm.~4.5, Remark~D.1]{yang2026lifted}.

\paragraph{ALBUM.}
Approximate Lifted Bellman Unconstrained Minimization implements LBLP with neural $Q_\theta,V_\phi$ by minimizing the hinge relaxation
\begin{equation}
  \label{eq:album}
  \begin{aligned}
  \mathcal L_{\mathrm{ALBUM}}=\;&\mathbb{E}_\mathcal{D}[\omega_{Q} Q+\omega_{V} V]+\lambda_{\mathrm{B}}\,\mathbb{E}_\mathcal{D}[g_{\mathrm{B}}]_{+}+\lambda_{\mathrm{E}}\,\mathbb{E}_\mathcal{D}[g_{\mathrm{E}}]_{+}\\
  &+\lambda_{\mathrm{K}}\big(\mathbb{E}_\mathcal{D}[\operatorname{sg}[\hat y_K]-Q]_{+}+\mathbb{E}_\mathcal{D}[\operatorname{sg}[\hat y_K]-V]_{+}\big),
  \end{aligned}
\end{equation}
with fixed coefficients $(\lambda_{\mathrm{B}},\lambda_{\mathrm{K}},\lambda_{\mathrm{E}})=(2.5,1,2.5)$ shared by every row and every environment, and with the $K$-step target detached by stop-gradient. Two results of \citet{yang2026lifted} matter here, and both are conditions on coefficients fixed in advance. Their Proposition~4.7 (exactness): in the tabular case the relaxation has the minimizer $(Q^*_\mathcal{D},V^*_\mathcal{D})$ when $\lambda_{\mathrm{B}}$ and $\lambda_{\mathrm{E}}$ exceed the largest optimal dual variables of the one-step program. This holds for every $\lambda_{\mathrm{K}}\ge0$ and $K$, but the thresholds depend on the dataset and are not known beforehand. Their Proposition~5.1 (bounded update): the detached update does not lower $Q$ and $V$ without bound along a uniform shift only if $\lambda_{\mathrm{B}}(1-\gamma)+2\lambda_{\mathrm{K}}\ge\omega_{Q}+\omega_{V}$. ALBUM's default coefficients satisfy this inequality with room to spare. Our AL-LBLP learns the multipliers instead of fixing them, and the second condition then holds with equality at every stationary point (\cref{sec:theory}).
\vspace{-3pt}
\section{AL-LBLP: Augmented Lagrangian for Lifted Bellman Linear Programming}
\label{sec:method}

\subsection{Method of multipliers}
\label{sec:alm}
We rewrite the constraint rows of \cref{eq:lblp} as $y^F_i\le0$ for family $F\in\{\mathrm B,\mathrm{KQ},\mathrm{KV},\mathrm E\}$ and window $i$, with $y>0$ a violation. The Lagrangian of LBLP with per-row multipliers $\mu^F_i\ge0$ is $\Lambda(Q,V,\mu)=\mathbb{E}_\mathcal{D}[\omega_{Q} Q+\omega_{V} V]+\sum_F\mathbb{E}_i[\mu^F_i\,y^F_i]$. Since $\sup_{\mu\in[0,\lambda]}\mu y=\lambda[y]_{+}$, the ALBUM loss \eqref{eq:album} is $\sup_\mu\Lambda$ over multipliers confined to the box $[0,\lambda_F]$. AL-LBLP keeps this Lagrangian and learns the multipliers.

The augmented Lagrangian of an inequality constraint $y\le0$ with multiplier $\mu\ge0$ and penalty $\rho>0$ is \citep{rockafellar1973dual,bertsekas1982constrained}
\begin{equation}
  \label{eq:psi}
  \begin{aligned}
  \psi_\rho(y;\mu)&=\frac{1}{2\rho}\Big([\mu+\rho y]_{+}^2-\mu^2\Big)\\
  &=\begin{cases} y\,(\mu+\tfrac{\rho}{2}y) & \text{if } \mu+\rho y>0,\\[2pt] -\mu^2/(2\rho) & \text{otherwise},\end{cases}
  \end{aligned}
\end{equation}
whose derivative in $y$ is $[\mu+\rho y]_{+}$. The AL-LBLP critic loss is
\begin{equation}
  \label{eq:loss}
  \begin{aligned}
  \mathcal L_\rho(\theta,\phi;\mu)=\;&\mathbb{E}_\mathcal{D}[\omega_{Q} Q_\theta+\omega_{V} V_\phi]
  +\sum_F\mathbb{E}_i\big[\psi_\rho\big(y^F_i;\operatorname{sg}[\mu^F_i]\big)\big],
  \end{aligned}
\end{equation}
where, as in ALBUM, the $K$-step target $\operatorname{sg}[\hat y_K]$ is detached and $V(s')$ in the one-step row carries gradient. Compared with ALBUM, the gradient of \cref{eq:loss} replaces the fixed weight $\lambda_F$ of each violated row by the \emph{effective multiplier} $[\mu^F_i+\rho y^F_i]_{+}$: a row with $\mu+\rho y\le0$ gets no penalty gradient, and a violated row's weight grows with its violation. After each primal step the multipliers move by projected dual ascent on the pre-update violations,
\begin{equation}
  \label{eq:dual}
  \mu^F_i\leftarrow[\mu^F_i+\rho\,\operatorname{sg}[y^F_i]]_{+},
\end{equation}
which accumulates persistent violations into $\mu$. As $\rho\to0$ the loss tends to the Lagrangian with frozen multipliers and the dual step vanishes. As $\rho\to\infty$ it tends to a quadratic penalty. $\rho$ is therefore both the step size of \cref{eq:dual} and the curvature of the penalty on violated rows. It is the only coefficient of the relaxation.

\paragraph{Multiplier parameterization.}
The multipliers can be a table with one entry per row, a scalar per family, or a network $\mu_\varphi(s,a)$. We use one scalar per family, $\mu_{\mathrm{B}},\mu_{\mathrm{K}},\mu_{\mathrm{E}}$, with $\mu_{\mathrm{K}}$ shared by the two rollout rows as ALBUM shares $\lambda_{\mathrm{K}}$. The scalar form approximates the per-row method. Its dual fixed point is an averaged form of complementary slackness (\cref{rem:scalar}). The dual step \eqref{eq:dual} regresses $\mu^F$ onto the batch mean of its per-row targets $[\mu^F+\rho\,\bar y^F_i]_{+}$ with its own Adam optimizer. Here $\bar y^F_i=y^F_i$ for the one-step and epigraph rows and $\bar y^{\mathrm K}_i=\tfrac12(y^{\mathrm{KQ}}_i+y^{\mathrm{KV}}_i)$ for the shared rollout multiplier. $\mu^F=\mathrm{softplus}(\zeta^F)$ keeps the multipliers non-negative, the optimizer acts on $\zeta^F$, and the multipliers start at $10^{-3}$. \Cref{alg:allblp} summarizes one training step.

\begin{algorithm}[tb]
  \caption{AL-LBLP (scalar multipliers)}
  \label{alg:allblp}
  \begin{algorithmic}
    \STATE {\bfseries Input:} dataset $\mathcal{D}$ of $K$-step windows, weights $\omega_{Q},\omega_{V}$, penalty $\rho$, BC coefficient $\alpha$
    \STATE Initialize $Q_\theta$, $V_\phi$, actor $\pi_\eta$, multipliers $\mu_{\mathrm{B}},\mu_{\mathrm{K}},\mu_{\mathrm{E}}\approx0$
    \REPEAT
    \STATE Sample a batch of windows $\{(s_i,a_i,r_i,s_i',\hat y_{K,i})\}$
    \STATE Violations: $y^{\mathrm B}_i=r_i+\gamma V_\phi(s_i')-Q_\theta(s_i,a_i)$, $y^{\mathrm E}_i=Q_\theta(s_i,a_i)-V_\phi(s_i)$,
    \STATE $y^{\mathrm{KQ}}_i=\operatorname{sg}[\hat y_{K,i}]-Q_\theta(s_i,a_i)$, $y^{\mathrm{KV}}_i=\operatorname{sg}[\hat y_{K,i}]-V_\phi(s_i)$
    \STATE Primal: one Adam step on $\mathcal L_\rho(\theta,\phi;\operatorname{sg}[\mu])$ of \cref{eq:loss}
    \STATE Dual: for each family $F$, one Adam step of $\mu^F=\mathrm{softplus}(\zeta^F)$ toward $\mathbb{E}_i\big[[\mu^F+\rho\,\bar y^F_i]_{+}\big]$, $\bar y^{\mathrm K}_i=\tfrac12(y^{\mathrm{KQ}}_i+y^{\mathrm{KV}}_i)$
    \STATE Actor: one Adam step on $\mathcal L_{\mathrm{actor}}(\pi_\eta)$ of \cref{eq:actor}
    \UNTIL{the step budget is spent}
  \end{algorithmic}
\end{algorithm}

\subsection{Policy extraction}
\label{sec:actor}
The critic is trained only at dataset actions and never evaluates the policy, so any extraction method can be attached to it. We report three. The default is the DDPG+BC objective used by ALBUM, and the policy-improvement term is normalized by the batch scale of the advantage $A(s,a)=Q(s,a)-V(s)$,
\begin{equation}
  \label{eq:actor}
  \mathcal L_{\mathrm{actor}}(\pi)=-\frac{\mathbb{E}_{s}\,[A(s,\pi(s))]}{\operatorname{sg}\,[\mathbb{E}_{s}[|A(s,\pi(s))|]]}-\alpha\,\mathbb{E}_{(s,a)\sim\mathcal{D}}\log\pi(a\mid s).
\end{equation}
The second term is the behavior-cloning (BC) term, and $\pi(s)$ denotes the mean of the Gaussian policy $\pi(\cdot\mid s)$.
We propose this normalization in place of the $|Q|$ normalization of TD3+BC \citep{fujimoto2021minimalist}, which ALBUM also uses. The LP critic provides $V$, so $|A|$ is available directly. The $|A|$-normalized update is invariant to common shifts $\delta(s)$ of $Q$ and $V$ and, in continuing tasks, to affine maps $r\mapsto cr+b$ ($c>0$) of the reward. The $|Q|$ normalization is invariant only when $b=0$ (\cref{prop:invariance}). \Cref{sec:norm} measures the effect on the same critic. The second actor is the one-step flow actor of FQL \citep{park2025flow}, whose policy-improvement term we normalize by $\mathbb{E}_s|A(s,\pi(s))|$ in the same way. The third, best-of-$N$, draws $N$ samples from a behavior-cloning flow and executes the one with the largest $Q$. None of these three actors changes the critic.

\begin{table*}[t]
  \centering
  \caption{Success rate (\%) on the 50 state-based tasks: five-task mean, mean $\pm$ std over 8 seeds (IFQL, FAC, DROL: mean only); AL-LBLP and ALBUM: the final evaluation at 1M, baselines: their own rule (\cref{app:details}). Bold: within 95\% of the row best.}
  \label{tab:main}
    \begin{small}
    \resizebox{\textwidth}{!}{%
    \setlength{\tabcolsep}{4pt}
        \begin{tabular}{lcccccccccc}
          \toprule
          & \multicolumn{4}{c}{Gaussian policy} & \multicolumn{4}{c}{Flow policy} & \multicolumn{2}{c}{Best-of-$N$} \\
          \cmidrule(lr){2-5}\cmidrule(lr){6-9}\cmidrule(lr){10-11}
          Environment & IQL & ReBRAC & ALBUM & \textbf{AL-LBLP} & FQL & DROL & FAC & \textbf{AL-LBLP} & IFQL & \textbf{AL-LBLP} \\
          \midrule
          antmaze-large       & 53 $\pm$3 & 81 $\pm$5 & \textbf{89.8} $\pm$1.0 & \textbf{92.2} $\pm$1.5 & 79 $\pm$3 & \textbf{89} & \textbf{92.6} & \textbf{89.0} $\pm$2.9 & 28.0 & 76.8 $\pm$4.3 \\
          antmaze-giant       & 4 $\pm$1 & 26 $\pm$8 & 28.1 $\pm$5.1 & \textbf{47.5} $\pm$7.4 & 9 $\pm$6 & 45 & 23.0 & 29.4 $\pm$4.0 & 2.6 & 34.9 $\pm$3.4 \\
          humanoidmaze-medium & 33 $\pm$2 & 22 $\pm$8 & 35.5 $\pm$1.8 & 85.6 $\pm$2.7 & 58 $\pm$5 & 68 & 75.6 & 78.5 $\pm$3.8 & 60.4 & \textbf{96.1} $\pm$1.2 \\
          humanoidmaze-large  & 2 $\pm$1 & 2 $\pm$1 & 1.6 $\pm$0.5 & 17.4 $\pm$1.0 & 4 $\pm$2 & 11 & 8.3 & 12.4 $\pm$3.5 & 11.0 & \textbf{47.7} $\pm$4.3 \\
          antsoccer-arena     & 8 $\pm$2 & 0 $\pm$0 & 36.1 $\pm$3.8 & 48.5 $\pm$4.5 & 60 $\pm$2 & \textbf{65} & \textbf{67.7} & 56.5 $\pm$5.9 & 33.2 & 56.1 $\pm$4.4 \\
          cube-single         & 83 $\pm$3 & 91 $\pm$2 & 83.2 $\pm$1.7 & 89.2 $\pm$2.7 & \textbf{96} $\pm$1 & \textbf{98} & \textbf{98.8} & 91.3 $\pm$1.1 & 79.2 & 89.5 $\pm$2.2 \\
          cube-double         & 7 $\pm$1 & 12 $\pm$1 & 13.3 $\pm$2.8 & 16.4 $\pm$2.4 & 29 $\pm$2 & 26 & \textbf{33.1} & 17.5 $\pm$3.6 & 14.0 & 20.2 $\pm$3.6 \\
          scene               & 28 $\pm$1 & 41 $\pm$3 & 64.4 $\pm$5.5 & 58.5 $\pm$2.5 & 56 $\pm$2 & 58 & \textbf{71.3} & 62.3 $\pm$3.4 & 30.4 & 63.0 $\pm$1.1 \\
          puzzle-3x3          & 9 $\pm$1 & 21 $\pm$1 & 61.2 $\pm$6.4 & 85.8 $\pm$6.6 & 30 $\pm$1 & 41 & \textbf{100.0} & \textbf{97.6} $\pm$1.3 & 19.0 & \textbf{96.1} $\pm$3.1 \\
          puzzle-4x4          & 7 $\pm$1 & 14 $\pm$1 & 20.6 $\pm$2.5 & 17.3 $\pm$6.5 & 17 $\pm$2 & 25 & \textbf{32.3} & 22.4 $\pm$4.9 & 25.2 & 21.0 $\pm$2.3 \\
          \midrule
          Average                      & 23.4 & 31.0 & 43.4 & 55.8 & 43.8 & 52.6 & \textbf{60.3} & 55.7 & 30.3 & \textbf{60.1} \\
          \bottomrule
        \end{tabular}
    }
    \end{small}
\vspace{-5pt}
\end{table*}

\subsection{Properties}
\label{sec:theory}
We consider convex instances of LBLP, in which $Q$ and $V$ are free variables on $\mathcal{M}_\mathcal{D}$ or linear in fixed features, with one multiplier per row and no stop-gradient. There the fixed points of primal minimization and the dual update \eqref{eq:dual} are the KKT pairs of LBLP, for every $\rho>0$. In the tabular case every fixed point satisfies $(Q,V)=(Q^*_\mathcal{D},V^*_\mathcal{D})$. When the fixed coefficients of ALBUM exceed the dual thresholds of its Proposition~4.7, the two methods have the same solution (\cref{cor:album}). AL-LBLP does not use the thresholds, and below them its fixed points are still the LBLP solution.

Beyond the convex case, let $Q_\theta$ and $V_\phi$ carry additive output biases. At every stationary point of \cref{eq:loss} the effective multipliers $\mu^{F,\mathrm{eff}}_i=[\mu^F_i+\rho y^F_i]_{+}$ satisfy the mass identity:
\begin{equation}
  \label{eq:mass}
  \mathbb{E}_i\big[(1-\gamma m_i)\,\mu^{\mathrm{B,eff}}_i+\mu^{\mathrm{KQ,eff}}_i+\mu^{\mathrm{KV,eff}}_i\big]=\omega_{Q}+\omega_{V} ,
\end{equation}
with $m_i$ the terminal mask. For constant weights and $m\equiv1$ the left-hand side is $\lambda_{\mathrm{B}}(1-\gamma)+2\lambda_{\mathrm{K}}$. The bounded-update condition of ALBUM therefore holds with equality for the weights that AL-LBLP applies, by the optimization itself. At 1M steps the left-hand side is between 0.148 and 0.154 against $\omega_{Q}+\omega_{V}=0.15$ on all ten environments (\cref{app:mass}). Formal statements and proofs are in \cref{app:proofs}.

\section{Experiments}
\label{sec:experiments}

\paragraph{Protocol.}
We follow the evaluation protocol of ALBUM \citep{yang2026lifted}: the 50 state-based single-task OGBench tasks \citep{park2024ogbench} (10 environments $\times$ 5 tasks), 8 seeds, and the success rate (\%) of the final evaluation after 1M gradient steps with 50 episodes. Every environment uses $\gamma=0.995$, $K=10$, a single $Q$ and $V$, and no target network. We select $(\rho,\alpha)$ on the default task of each environment with 3 seeds and reuse it on the other four tasks (\cref{tab:side-picks}). Baselines are not re-run: IQL, ALBUM, ReBRAC, FQL, IFQL, FAC, and DROL are the scores reported in their papers. All other settings, including the grids, the baseline rules, and the fixed hyperparameters, are in \cref{app:details}.

\subsection{Main results}
\label{sec:main}

AL-LBLP reaches its results with a single critic and no target network, and the same critic serves all three policies. \Cref{tab:main} groups the methods by the policy they execute. The Gaussian group holds IQL, ReBRAC, ALBUM, and AL-LBLP. The flow group holds FQL, FAC, DROL, and AL-LBLP with a variant of the one-step flow actor of FQL. The best-of-$N$ group holds IFQL and AL-LBLP, both sampling from a behavior-cloning flow.

With the Gaussian actor AL-LBLP averages 55.8, against 43.4 for ALBUM, which solves the same program with fixed hinge coefficients, and 43.8 for FQL. It outperforms every baseline on \textit{antmaze-giant} (47.5), \textit{humanoidmaze-medium} (85.6), and \textit{humanoidmaze-large} (17.4), and matches FAC on \textit{antmaze-large}. Relative to ALBUM the gains concentrate on the navigation environments (\textit{humanoidmaze-medium} 35.5 to 85.6, \textit{humanoidmaze-large} 1.6 to 17.4, \textit{antmaze-giant} 28.1 to 47.5) and on \textit{puzzle-3x3} (61.2 to 85.8). ALBUM remains ahead on \textit{scene} and \textit{puzzle-4x4}.

The critic does not depend on the actor's type or hyperparameters. Therefore, the flow-policy and best-of-$N$ columns attach other actors to critics of the same program as for the Gaussian policy. Each has its own $(\rho,\alpha_{\mathrm{flow}})$ or $(\rho,N)$ selected from six combinations, against nine for the Gaussian policy. AL-LBLP has the highest average among the Gaussian policies (55.8) and among the best-of-$N$ methods (60.1 against 30.3 for IFQL). Among the flow policies it is second (55.7), after 60.3 for FAC and ahead of 52.6 for DROL and 43.8 for FQL.

Best-of-$N$ is the strongest of the three, close to FAC on average and ahead of it on \textit{antmaze-giant} and both \textit{humanoidmaze} environments. Against the Gaussian policy it gains most on \textit{humanoidmaze-large} (47.7 against 17.4), \textit{humanoidmaze-medium} (96.1 against 85.6), and \textit{puzzle-3x3} (96.1 against 85.8), and is lower on \textit{antmaze-large} (76.8 against 92.2) and \textit{antmaze-giant} (34.9 against 47.5). Per-task results are in \cref{app:pertask}. On the five pixel-based manipulation environments AL-LBLP (Gaussian policy) averages 64.7 against 65.4 for FQL and 76.0 for FAC (\cref{app:visual}).
\vspace{+4pt}
\subsection{Long-horizon tasks and large batch size}
\label{sec:longhorizon_batch}

On \textit{antmaze-giant} and \textit{humanoidmaze-giant} we raise the rollout length to $K=50$ and the discount to $\gamma=0.999$. The error of the idealized ALBUM scheme contracts at rate $\gamma^K$ where the $K$-step path is optimal (Theorem~4.5 of \citet{yang2026lifted}). Raising $\gamma$ with $K$ keeps this rate at the value of the shared protocol ($0.995^{10}=0.9511$, $0.999^{50}=0.9512$). \cref{tab:longhorizon} reports 8 seeds per task at batch sizes 256 to 8192, with the $(\rho,\alpha)$ selected at batch 256; rows marked $^\dagger$ instead re-select $(\rho,\alpha)$ for each batch size (\textit{i.e.,} 2048, 4096 and 8192). The selected $(\rho,\alpha)$ are in \cref{app:details}, and the $K=10$ settings are those of \cref{sec:main}.

\begin{figure*}[!t]
\vspace{-8pt}
  \centering
  \includegraphics[width=\textwidth]{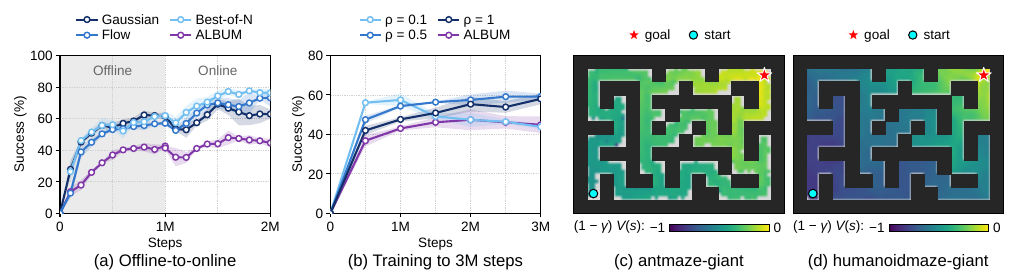}
  \caption{(a) Offline-to-online fine-tuning of the three AL-LBLP policies and ALBUM: mean success (\%) over the 10 default tasks during offline training (grey) and online fine-tuning; mean $\pm$ std over 3 seeds. (b) Training to 3M steps with $\rho\in\{0.1,0.5,1\}$ ($\alpha$ of \cref{fig:rho-norm}(a)) and ALBUM: mean success (\%) over the 10 default tasks, evaluated every 0.5M steps; mean $\pm$ std over 3 seeds. (c), (d) Learned value $(1-\gamma)V(s)$ of AL-LBLP ($K=50$, $\gamma=0.999$, batch 4096, $(\rho,\alpha)$ of the $^\dagger$ rows of \cref{tab:longhorizon}) on task 1 (details in \cref{app:longhorizon}).}
  \label{fig:combined_results}
\vspace{-10pt}
\end{figure*}

\subsubsection{Long-horizon navigation tasks}
\label{sec:longhorizon}

AL-LBLP performs especially well on long-horizon navigation tasks. With 1M offline steps it reaches 81.5 $\pm$5.9 on \textit{humanoidmaze-giant} (batch 4096) and 79.8 $\pm$2.5 on \textit{antmaze-giant} (batch 4096), five-task means over 8 seeds. Both are higher than the best scores reported by LQL \citep{abraham2026long}, 65.4 on \textit{antmaze-giant} and 75.7 on \textit{humanoidmaze-giant} (4 seeds). LQL reaches them after 2M training steps, at least 1M of them online, whereas AL-LBLP uses 1M offline steps only (details in \cref{app:longhorizon}).

On \textit{humanoidmaze-giant}, AL-LBLP also matches Transitive RL (TRL) \citep{park2026transitive}, which reports 79 $\pm$2 (4 seeds) in the goal-conditioned version of the task. TRL trains larger networks ($4\times1024$) on a dataset of 1B transitions \citep{park2025horizon} and relabels goals with future states of the same trajectory, so many of its training goals are close to the current state. In contrast, AL-LBLP uses the default dataset of 4M transitions, $4\times512$ networks, and the single-task reward, which is $-1$ at every step until the one evaluation goal is reached. Its critic thus propagates the signal of a single goal over episodes of up to 4000 steps, whereas TRL learns from many nearby goals and sub-goals.

Raising $K$ to 50 together with $\gamma$ to 0.999 accounts for a large part of the gain. On \textit{antmaze-giant} (batch 256) the five-task mean rises from 47.5 $\pm$7.4 to 64.6 $\pm$3.8. ALBUM does not benefit from the same change. With $K=50$, $\gamma=0.999$, batch 256, and $\alpha$ selected by the same protocol, it reaches 29.5 $\pm$2.8 on \textit{antmaze-giant}, only slightly above 28.1 $\pm$5.1 at $\gamma=0.995$, $K=10$, and 1.2 $\pm$0.5 on \textit{humanoidmaze-giant} (five tasks, 8 seeds). On both environments it falls far behind AL-LBLP at the same batch size (64.6 and 71.0).

\begin{table}[t]
  \centering
  \caption{Batch size on the long-horizon navigation tasks: five-task success (\%) at 1M, mean $\pm$ std over 8 seeds. $K=10$: $\gamma=0.995$; $K=50$: $\gamma=0.999$. $^\dagger$: $(\rho,\alpha)$ re-selected per batch size. Bold: within 95\% of the row best.}
  \label{tab:longhorizon}
    \begin{small}
      \resizebox{\columnwidth}{!}{%
      \begin{tabular}{lcccc}
        \toprule
        & \multicolumn{4}{c}{Batch size} \\
        \cmidrule(lr){2-5}
        Environment & 256 & 2048 & 4096 & 8192 \\
        \midrule
        antmaze-giant ($K=10$) & 47.5 $\pm$7.4 & \textbf{56.1} $\pm$4.0 & \textbf{53.7} $\pm$3.9 & 52.2 $\pm$4.5 \\
        antmaze-giant ($K=50$) & 64.6 $\pm$3.8 & \textbf{70.8} $\pm$3.7 & 66.0 $\pm$2.2 & 59.5 $\pm$3.6 \\
        antmaze-giant ($K=50$)$^\dagger$ & 64.6 $\pm$3.8 & \textbf{78.4} $\pm$3.0 & \textbf{79.8} $\pm$2.5 & \textbf{77.4} $\pm$2.5 \\
        \midrule
        humanoidmaze-large ($K=10$) & 17.4 $\pm$1.0 & 32.9 $\pm$7.0 & \textbf{39.9} $\pm$6.5 & 33.1 $\pm$7.5 \\
        humanoidmaze-giant ($K=50$) & 71.0 $\pm$3.5 & \textbf{77.8} $\pm$7.7 & \textbf{79.7} $\pm$4.3 & \textbf{79.3} $\pm$5.7 \\
        humanoidmaze-giant ($K=50$)$^\dagger$ & 71.0 $\pm$3.5 & \textbf{80.8} $\pm$3.6 & \textbf{81.5} $\pm$5.9 & \textbf{80.8} $\pm$3.2 \\
        \bottomrule
      \end{tabular}
      }
    \end{small}
\vspace{-15pt}
\end{table}

\subsubsection{Large batch size}
\citet{park2026scalable} reports that TD learning with a large fixed batch size degrades performance across network sizes. Under our shared protocol, the TD-based FQL and FAC collapse when only their batch size grows: both score 0.0 on both environments (\textit{i.e.,} \textit{antmaze-giant} and \textit{humanoidmaze-large}) at batch 4096 and at most 4.9 at batch 2048 (\cref{tab:batch-fql}). In contrast, AL-LBLP does not collapse when only its batch size grows, and its performance even improves: with the $(\rho, \alpha)$ of batch 256, every value at batch 2048, 4096, and 8192 is above its row's batch-256 value, except \textit{antmaze-giant} ($K=50$) at 8192, which keeps 92\% of its batch-256 success (\cref{tab:longhorizon}).

Re-selecting $(\rho, \alpha)$ at each batch size within the same budget removes the drop of \textit{antmaze-giant} ($K=50$): the $^\dagger$ row reaches 78.4, 79.8, and 77.4 at batch 2048, 4096, and 8192, against 64.6 at batch 256, and \textit{humanoidmaze-giant}$^\dagger$ reaches 81.5, the best in \cref{tab:longhorizon}.
A plausible reason is that a larger batch gives a better estimate of the objective and of each constraint family's violation.

\begin{figure*}[!t]
\vspace{-5pt}
  \centering
  \includegraphics[width=\textwidth]{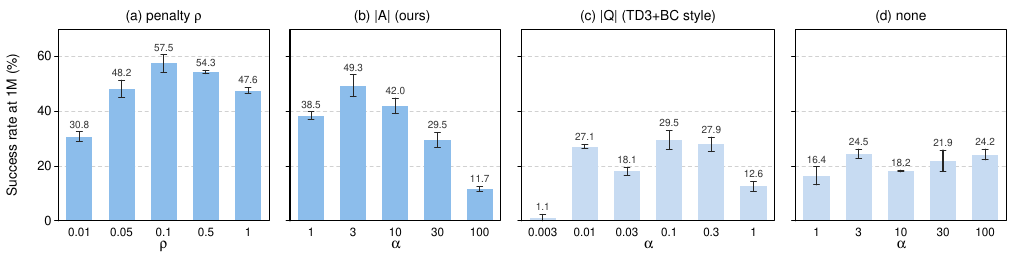}
  \caption{Mean success (\%) over the 10 default tasks at 1M, mean $\pm$ std over 3 seeds. (a) Penalty $\rho$, $\alpha$ selected per environment. (b)--(d) Actors normalizing the $Q$ term by $|A|$ (ours), by $|Q|$, or not at all, on the same AL-LBLP critic with $\rho$ fixed at 0.1.}
  \label{fig:rho-norm}
\vspace{-8pt}
\end{figure*}

\subsection{Offline-to-online fine-tuning}
\label{sec:off2on}

On average, all policies improve online with the same critic loss (\cref{fig:combined_results}(a)). We follow the offline-to-online protocol of FQL \citep{park2025flow} with the offline hyperparameters of each policy and 3 seeds. Each run restores the 1M offline checkpoint of its default task and trains for 1M online steps (\cref{app:details}). Over the ten default tasks the flow policy rises from 57.1 to 73.1 and best-of-$N$ from 61.9 to 76.3. The Gaussian policy moves from 61.5 to 62.9. The largest gains are on \textit{antmaze-giant}, where the Gaussian policy goes from 34.7 to 94.7 and best-of-$N$ from 0.0 to 93.3. Best-of-$N$ also goes from 59.3 to 86.7 on \textit{humanoidmaze-large}, and the flow policy from 24.0 to 97.3 on \textit{puzzle-4x4}. The Gaussian policy declines on \textit{puzzle-3x3} (97.3 to 33.3) and \textit{scene} (96.0 to 74.7). Per-environment results are in \cref{app:off2on}.
\vspace{+8pt}
\subsection{Ablations}
\label{sec:ablations}

\subsubsection{The penalty parameter $\rho$}
\label{sec:rho}

$\rho$ is the step size of \cref{eq:dual} and the stiffness of the penalty on violated rows. With a small $\rho$ the penalty grows slowly with the violation, so the constraints bind loosely. A large $\rho$ enforces every violated row stiffly from the start. \Cref{fig:rho-norm}(a) varies $\rho$ from 0.01 to 1 and re-selects $\alpha$ per environment for each $\rho$. At 1M the 10-environment mean peaks at $\rho=0.1$ (57.5) and falls on both sides, to 30.8 at $\rho=0.01$ and 47.6 at $\rho=1$.

\Cref{fig:combined_results}(b) trains $\rho\in\{0.1,0.5,1\}$ to 3M steps with the $\alpha$ selected at 1M (per environment in \cref{tab:rho}). A small $\rho$ learns fast early: at 0.5M the mean is 56.1 for $\rho=0.1$, 47.5 for $\rho=0.5$, and 42.0 for $\rho=1$. With longer training $\rho=0.1$ degrades, to 43.9 at 3M, and collapses on \textit{scene}, from 91.3 at 1M to 8.7. In contrast, a large $\rho$ is slow early but learns stably. $\rho=0.5$ goes from 54.3 at 1M to 59.2 at 3M, and $\rho=1$ from 47.6 to 58.0 (on \textit{humanoidmaze-medium} from 48.7 to 94.0). The curves cross between 1M and 1.5M. This is a trade-off between early speed and long-run stability: a small $\rho$ learns fast under loose constraints, and a large $\rho$ stays stable under tight ones.

\Cref{tab:rho-speed} reports the relative size of the effective multipliers $w=[\mu+\rho y_i]_{+}$ of the $K$-step rows, $w^{K}_{\max}/\bar w^{K}$ (batch maximum over batch mean in the first 100k steps), and the srank of the critic \citep{kumar2021implicit}. Since $w$ grows with the violation $y_i$ at rate $\rho$, a larger $\rho$ concentrates the update on a few badly violated rows: $w^{K}_{\max}/\bar w^{K}$ is larger for $\rho=1$ than for $\rho=0.1$ in all 10 environments. A small $\rho$ instead keeps a higher-rank representation: the srank is higher for $\rho=0.1$ than for $\rho=1$ in all 10 environments. \citet{park2025pretraining} relate a higher rank of the $Q$-network's features to a smaller error bound and faster convergence in offline RL. Consistent with this, across the 30 ($\rho$, environment) cells the srank is rank-correlated with the success at 0.5M (Spearman 0.53 after centering each environment).

This trade-off between early speed and long-run stability matches \cref{tab:rho-speed,tab:rho-viol}: a small $\rho$ keeps a high srank but leaves larger violations, while a large $\rho$ shrinks the srank but keeps the constraints tight.

\begin{table}[t]
  \centering
    \caption{Concentration of the $K$-step effective multipliers (first 100k steps) and critic srank (0.5M) on the 10 default tasks, mean $\pm$ std over 3 seeds. $w^K=[\mu+\rho y_i]_{+}$: effective multiplier of a $K$-step row; $w^{K}_{\max}/\bar w^{K}$: its batch maximum over its batch mean.}
    \label{tab:rho-speed}
    \begin{small}
    \begin{tabular}{lcc}
      \toprule
      $\rho$ & $w^{K}_{\max}/\bar w^{K}$ [$\downarrow$] & srank [$\uparrow$] \\
      \midrule
      0.1 & \textbf{26.7} $\pm$1.6 & \textbf{408} $\pm$1 \\
      0.5 & 65.7 $\pm$2.3 & 317 $\pm$7 \\
      1 & 93.7 $\pm$7.5 & 297 $\pm$8 \\
      \bottomrule
    \end{tabular}
    \end{small}
\vspace{-10pt}
\end{table}

\subsubsection{Actor normalization}
\label{sec:norm}
\cref{fig:rho-norm}(b) shows that the $|A|$ normalization of \cref{eq:actor} raises performance and narrows the useful range of $\alpha$. We attach three Gaussian actors to the same AL-LBLP critic with $\rho$ fixed at 0.1, the value with the highest 1M mean in \cref{fig:rho-norm}(a). The critic does not read the actor, and its logged $Q$, $V$, and multipliers are identical across the three actors for every environment and seed. The actors differ only in how the $Q$ term is normalized: by $\mathbb{E}_s|A(s,\pi(s))|$ as in \cref{eq:actor}, by $\mathbb{E}_s|Q(s,\pi(s))|$ as in TD3+BC \citep{fujimoto2021minimalist} and ALBUM \citep{yang2026lifted}, or not at all. With one $\alpha$ for every environment, the best 10-environment mean is 49.3 with $|A|$ ($\alpha=3$), 29.5 with $|Q|$ ($\alpha=0.1$), and 24.5 without normalization ($\alpha=3$) (\cref{fig:rho-norm}(b)--(d)).

With $|A|$ the best $\alpha$ varies little across environments: in \cref{fig:rho-norm}(b) the mean success peaks within $[1,10]$ on a grid up to 100, and the per-environment best $\alpha$ all lie in $[1,10]$, so three values $\{1,3,10\}$ are enough. In contrast, with $|Q|$ or without normalization it varies up to 100-fold, and ALBUM's selected values range from 0.003 to 0.3. The reason is that the two terms of \cref{eq:actor} have different scales across environments.

The behavior-cloning term depends only on the actions, which lie in $[-1,1]$ across all environments. Since the $Q$-term compares actions, its effective scale is determined by $\mathbb{E}_s|A|=\mathbb{E}_s|Q-V|$, which varies from 0.60 to 14.8 across environments (25-fold). Thus, without normalization, $\alpha$ must compensate for these scale differences. Normalizing by $\mathbb{E}_s|Q|$ (102 to 1065) does not resolve this issue, as $|Q|$ is dominated by the action-independent baseline $V(s)$. Indeed, the action differences in the $|Q|$-normalized term have scale $\mathbb{E}_s|A|/\mathbb{E}_s|Q|$, which still varies from $1/330$ to $1/22$ (15-fold) across environments.

In contrast, normalizing by $\mathbb{E}_s|A|$ gives them the scale $\mathbb{E}_s|A|/\mathbb{E}_s|A|=1$ in every environment, removing these scale differences and allowing a single $\alpha$ across environments. \Cref{prop:invariance} formalizes this distinction: the $|A|$-normalized update is invariant to affine reward transformations and common shifts of $Q$ and $V$, whereas the $|Q|$-normalized update changes under reward shifts and common shifts.

\subsubsection{Comparison with ALBUM}
\label{sec:album-cmp}

\begin{table}[t]
  \centering
  \caption{$K$-step violation $y^{\mathrm{KQ}}$ at 1M steps on the giant mazes (five tasks), mean $\pm$ std over 8 seeds; $y^{\mathrm{KQ}}>0$: violation, $y^{\mathrm{KQ}}<0$: slack. Goal: windows reaching the goal. $^\dagger$: batch 4096.}
  \label{tab:goal-rows}
    \begin{small}
      \resizebox{\columnwidth}{!}{%
        \begin{tabular}{llcccc}
          \toprule
          & & \multicolumn{2}{c}{$\mathbb{E}[y^{\mathrm{KQ}}]$} & \multicolumn{2}{c}{$P(y^{\mathrm{KQ}}>0)$ (\%)} \\
          \cmidrule(lr){3-4}\cmidrule(lr){5-6}
          Environment & Method & Goal & All & Goal & All \\
          \midrule
          \multirow{3}{*}{antmaze-giant} & ALBUM & 356.72 $\pm$72.88 & $-$13.18 $\pm$2.88 & 47.8 $\pm$5.6 & 21.4 $\pm$15.7 \\
           & AL-LBLP & 0.02 $\pm$0.59 & $-$32.61 $\pm$0.07 & 45.1 $\pm$5.1 & 21.7 $\pm$2.6 \\
           & AL-LBLP$^\dagger$ & $-$1.82 $\pm$0.29 & $-$33.07 $\pm$0.12 & 28.3 $\pm$4.9 & 16.6 $\pm$3.4 \\
          \midrule
          \multirow{3}{*}{humanoidmaze-giant} & ALBUM & 585.92 $\pm$1.01 & $-$6.71 $\pm$1.33 & 72.4 $\pm$3.9 & 11.1 $\pm$10.5 \\
           & AL-LBLP & 1.69 $\pm$0.66 & $-$20.36 $\pm$0.19 & 61.5 $\pm$3.8 & 21.0 $\pm$0.7 \\
           & AL-LBLP$^\dagger$ & $-$1.20 $\pm$0.44 & $-$21.50 $\pm$0.14 & 43.4 $\pm$3.5 & 16.8 $\pm$0.4 \\
          \bottomrule
        \end{tabular}
      }
    \end{small}
\vspace{-10pt}
\end{table}

ALBUM and AL-LBLP solve the same program but enforce its constraints differently: fixed hinge coefficients against learned multipliers with a quadratic term. On the default tasks ALBUM violates its rows least often, but its hinge applies the same gradient $\lambda$ to every violated row, while the effective multiplier $[\mu+\rho y]_{+}$ of AL-LBLP grows with the violation. ALBUM thus leaves larger $K$-step violations (5.17 against 0.71 at $\rho=1$, \cref{tab:rho-viol}).

\Cref{tab:goal-rows} shows where this matters. Over all windows of the giant mazes, ALBUM violates the $K$-step rows no more often than AL-LBLP at batch 256. Only 0.3--0.6\% of the windows reach the goal, but there ALBUM violates the rows by hundreds, while AL-LBLP keeps them tight. Because these rows accelerate the propagation of value from the goal (Theorem~4.5 of \citet{yang2026lifted}), violating them affects the whole maze: ALBUM's $V$ stays at the floor on task 1 of both giant mazes, whereas AL-LBLP propagates the value from the goal (\cref{fig:value-heatmap}).

\Cref{tab:album-path} moves from AL-LBLP to ALBUM one component at a time. With $|Q|$ in place of $|A|$, the mean decreases from 47.6 to 46.3 at 1M steps. With fixed coefficients but the quadratic term, the $K$-step violations stay small (0.98), but the success drops to that of ALBUM (42.1). With learned multipliers and the hinge, the $K$-step violations are as large as in ALBUM (5.50), and the success falls to 30.0 at 1M. At the critic learning rate of ALBUM ($10^{-4}$), this row reaches 45.7 at 1M but falls to 33.8 at 3M, and its $K$-step violations stay large (3.32).

The results demonstrate that the quadratic term keeps the violations small, and the learned multipliers and advantage normalization are needed on top of it to reach the success of AL-LBLP.

\begin{table}[t]
  \centering
  \caption{From AL-LBLP to ALBUM, one component at a time: mean success (\%) over the 10 default tasks, mean $\pm$ std over 3 seeds; $K$-step violation size at 1M (\cref{tab:rho-viol}). Coeff.: learned $\mu$ or ALBUM's fixed $\lambda$. lr: critic learning rate.}
  \label{tab:album-path}
    \begin{small}
      \resizebox{\columnwidth}{!}{%
        \begin{tabular}{lccccccc}
          \toprule
          & Penalty & lr & Coeff. & Actor & 1M & 3M & $\mathbb{E}[y^{\mathrm{K}}\mid y^{\mathrm{K}}>0]$ \\
          \midrule
          AL-LBLP & quadratic & $3\times10^{-4}$ & $\mu$ & $|A|$ & 47.6 $\pm$1.1 & 58.0 $\pm$2.7 & 0.71 $\pm$0.02 \\
          $-$ $|A|$ & quadratic & $3\times10^{-4}$ & $\mu$ & $|Q|$ & 46.3 $\pm$1.7 & 51.7 $\pm$3.7 & 0.71 $\pm$0.02 \\
          $-$ multipliers & quadratic & $3\times10^{-4}$ & $\lambda$ & $|Q|$ & 42.1 $\pm$0.8 & 45.3 $\pm$2.5 & 0.98 $\pm$0.05 \\
          $-$ quadratic term & hinge & $3\times10^{-4}$ & $\mu$ & $|Q|$ & 30.0 $\pm$1.7 & 26.7 $\pm$2.7 & 5.50 $\pm$0.88 \\
          $-$ quadratic term & hinge & $10^{-4}$ & $\mu$ & $|Q|$ & 45.7 $\pm$5.5 & 33.8 $\pm$2.9 & 3.32 $\pm$0.32 \\
          \midrule
          ALBUM & hinge & $10^{-4}$ & $\lambda$ & $|Q|$ & 43.0 $\pm$1.0 & 44.9 $\pm$4.2 & 5.17 $\pm$2.53 \\
          \bottomrule
        \end{tabular}
      }
    \end{small}
\vspace{-8pt}
\end{table}
\vspace{5.5pt}
\section{Conclusion}
\label{sec:conclusion}
We solved the lifted Bellman linear program with the method of multipliers. The fixed hinge coefficients of ALBUM become multipliers learned by dual ascent. One penalty parameter $\rho$ sets both the dual step and the curvature on violated rows. In the convex case the fixed points are the KKT pairs of the program. At every stationary point the applied row weights balance the objective, which turns the bounded-update condition of ALBUM into an equality. The critic uses a single $Q$ and $V$ without a target network or ensemble and never queries the policy. We attached it to a Gaussian actor, for which we proposed an advantage-normalized DDPG+BC objective, to a flow actor, and to best-of-$N$ sampling. On the 50 state-based OGBench tasks the three policies average 55.8, 55.7, and 60.1. With longer rollouts the critic reaches 81.5 on \textit{humanoidmaze-giant} and 79.8 on \textit{antmaze-giant} with 8 seeds. The penalty parameter trades early speed against long-run stability. The gap to ALBUM lies in the $K$-step constraints at the goal, which AL-LBLP keeps tight and ALBUM does not.

\paragraph{Limitations.}
A small $\rho$ learns fast early but is less stable in long training: at $\rho=0.1$ the ten-environment mean falls from 56.1 at 0.5M to 43.9 at 3M (\cref{sec:rho}). AL-LBLP is weaker on manipulation than on navigation, in both the state-based and the pixel-based tasks (\cref{sec:main}, \cref{app:visual}). The scalar multipliers approximate the per-row method, and $\rho$ and $\alpha$ are still selected per environment.

\paragraph{Future work.}
Best-of-$N$ over a behavior-cloning flow is the strongest of our three policies, both offline (60.1, \cref{tab:main}) and after online fine-tuning (76.3, \cref{fig:combined_results}(a)). Many VLA models generate actions with a behavior-cloning flow, and the AL-LBLP critic is trained without querying the policy. Attaching an AL-LBLP critic to a VLA model by best-of-$N$ sampling is therefore our next step.

\section*{Impact Statement}
This paper presents work whose goal is to advance the field of Machine Learning. There are many potential societal consequences of our work, none which we feel must be specifically highlighted here.

\bibliography{refs}
\bibliographystyle{icml2026}

\newpage
\appendix
\onecolumn

\section{Proofs}
\label{app:proofs}
This appendix states and proves the properties summarized in \cref{sec:theory} and the invariance of the advantage-normalized actor of \cref{sec:actor}.

\paragraph{Setting and notation.}
In this appendix the primal variable is a finite vector $x$. In the tabular case its entries are the values of $Q$ on the dataset pairs and of $V$ on the dataset states. With linear features they are the weight vectors of $Q$ and $V$. The objective of \cref{eq:lblp} is linear, $f(x)=c^\top x$: the empirical weights $\omega_{Q}/n$ and $\omega_{V}/n$ on their coordinates, or their images under the feature maps. Every constraint row is affine, $y_j(x)=b_j-a_j^\top x$, with $y_j(x)>0$ a violation. We write $J$ for the set of rows, $y(x)=(y_j(x))_{j\in J}$, and $\mu=(\mu_j)_{j\in J}\ge0$ for one multiplier per row. Without the stop-gradient, the augmented Lagrangian of \cref{eq:loss} is
\begin{equation*}
  \mathcal L_\rho(x;\mu)=c^\top x+\sum_{j\in J}\psi_\rho\big(y_j(x);\mu_j\big),\qquad
  \psi_\rho(y;\mu)=\tfrac{1}{2\rho}\big([\mu+\rho y]_{+}^2-\mu^2\big),
\end{equation*}
and the dual update \eqref{eq:dual} is $\mu_j\leftarrow[\mu_j+\rho\,y_j(x)]_{+}$. We use two elementary facts about $\psi_\rho$.
\begin{itemize}
  \item[(F1)] For fixed $\mu\ge0$, $y\mapsto\psi_\rho(y;\mu)$ is convex and continuously differentiable with $\partial_y\psi_\rho(y;\mu)=[\mu+\rho y]_{+}$. On $\{\mu+\rho y>0\}$ it equals $\mu y+\tfrac{\rho}{2}y^2$, on the complement it is the constant $-\mu^2/(2\rho)$, and the two pieces and their derivatives agree at $\mu+\rho y=0$.
  \item[(F2)] $\psi_\rho(y;\mu)=\max_{\nu\ge0}\big[\nu y-\tfrac{1}{2\rho}(\nu-\mu)^2\big]$, attained at $\nu=[\mu+\rho y]_{+}$. Hence $\psi_\rho(\cdot;\mu)$ is the conjugate of a quadratic proximal term centered at $\mu$, which is the sense in which \cref{sec:related} calls the augmented Lagrangian a regularized dual \citep{rockafellar1973dual}.
\end{itemize}
Since each $y_j$ is affine, $\mathcal L_\rho(\cdot;\mu)$ is convex and $C^1$ in $x$ by (F1), with
\begin{equation}
  \label{eq:gradL}
  \nabla_x\mathcal L_\rho(x;\mu)=c-\sum_{j\in J}[\mu_j+\rho\,y_j(x)]_{+}\,a_j .
\end{equation}
A KKT pair of the LP $\min\{c^\top x: y(x)\le0\}$ is a pair $(x^*,\mu^*)$ with
\begin{equation}
  \label{eq:kkt}
  y_j(x^*)\le0,\qquad \mu^*_j\ge0,\qquad \mu^*_j\,y_j(x^*)=0\quad(j\in J),\qquad c-\sum_{j\in J}\mu^*_j a_j=0 .
\end{equation}
For a linear program with a finite optimum, $x^*$ is optimal if and only if there exists $\mu^*$ with which $(x^*,\mu^*)$ is a KKT pair. No constraint qualification is needed because the constraints are affine \citep[Thm.~28.2 and Cor.~28.3.1]{rockafellar1970convex}.

\paragraph{ALBUM in the dual.}
With the notation above, the ALBUM loss \eqref{eq:album} in the tabular case is $\max\{\Lambda(x,\mu):0\le\mu_j\le\lambda_{F(j)}/n\}$ with $\Lambda(x,\mu)=c^\top x+\sum_j\mu_j y_j(x)$ the ordinary Lagrangian, because $\max_{\mu\in[0,\lambda]}\mu y=\lambda[y]_{+}$. Minimizing it over $x$ is therefore the LP with the additional dual constraint $\mu_j\le\lambda_{F(j)}/n$ on every row. This is where the condition of Proposition~4.7 of \citet{yang2026lifted} comes from: when every optimal dual variable of the one-step program lies strictly inside its box, the added constraints are inactive and the relaxation is exact. AL-LBLP imposes no box on $\mu$, and the multipliers are determined by the dual update instead.

\subsection{Fixed points}
\begin{proposition}[Fixed points]
  \label{prop:fixed}
  Let LBLP \eqref{eq:lblp} be a convex instance with a finite optimum, and let $\rho>0$. A pair $\big((Q,V),\mu\big)$ with $\mu\ge0$ satisfies $(Q,V)\in\arg\min\mathcal L_\rho(\cdot;\mu)$ and $\mu=[\mu+\rho\,y(Q,V)]_{+}$ (a fixed point of the primal minimization and of the dual update \eqref{eq:dual}) if and only if it is a KKT pair of LBLP. In the deterministic tabular case $(Q,V)=(Q^*_\mathcal{D},V^*_\mathcal{D})$ at every fixed point. With linear features $(Q,V)$ is an optimal solution of the feature-restricted program.
\end{proposition}

\begin{proof}
Fix $\rho>0$. We show that the two fixed-point conditions together are equivalent to \cref{eq:kkt}.

\emph{Dual fixed point $\Leftrightarrow$ feasibility, dual feasibility, and complementary slackness.}
Suppose $\mu_j=[\mu_j+\rho\,y_j(x)]_{+}$ for every $j$. Then $\mu_j\ge0$. If $\mu_j>0$, the maximum is attained by the left argument, so $\mu_j=\mu_j+\rho\,y_j(x)$ and $y_j(x)=0$. If $\mu_j=0$, then $[\rho\,y_j(x)]_{+}=0$, so $y_j(x)\le0$. In both cases $y_j(x)\le0$ and $\mu_j\,y_j(x)=0$. Conversely, suppose $y_j(x)\le0$, $\mu_j\ge0$, and $\mu_j y_j(x)=0$. Then either $\mu_j=0$ and $[\rho y_j(x)]_{+}=0=\mu_j$, or $\mu_j>0$, $y_j(x)=0$, and $[\mu_j]_{+}=\mu_j$. Hence the dual fixed-point condition holds for every row if and only if the three conditions of \cref{eq:kkt} other than stationarity hold. In that case
\begin{equation}
  \label{eq:mueff_eq_mu}
  [\mu_j+\rho\,y_j(x)]_{+}=\mu_j\qquad\text{for every } j\in J .
\end{equation}

\emph{Primal minimizer under \cref{eq:mueff_eq_mu} $\Leftrightarrow$ stationarity.}
Because $\mathcal L_\rho(\cdot;\mu)$ is convex and differentiable, $x\in\arg\min\mathcal L_\rho(\cdot;\mu)$ if and only if $\nabla_x\mathcal L_\rho(x;\mu)=0$. Substituting \cref{eq:mueff_eq_mu} into \cref{eq:gradL} gives $\nabla_x\mathcal L_\rho(x;\mu)=c-\sum_j\mu_j a_j$. Under \cref{eq:mueff_eq_mu} the primal condition is therefore the last condition of \cref{eq:kkt}.

Combining the two steps, $(x,\mu)$ satisfies both fixed-point conditions if and only if it satisfies all of \cref{eq:kkt}, that is, if and only if it is a KKT pair. The argument used only $\rho>0$.

\emph{Identification of the primal point.}
The primal part of a KKT pair is an optimal solution of the LP. In the tabular case the LP is LBLP on $\mathcal{M}_\mathcal{D}$, whose minimizer is unique and equal to $(Q^*_\mathcal{D},V^*_\mathcal{D})$ \citep[Prop.~4.3]{yang2026lifted}. With linear features the LP is LBLP restricted to the span of the features, an approximate linear program in the sense of \citet{defarias2003linear}. Its optimal set need not be a singleton, and the conclusion is that $x$ belongs to it.
\end{proof}

\begin{remark}[Scalar multipliers]
  \label{rem:scalar}
  With one scalar per family the dual fixed point is $\mu^F=\mathbb{E}_i\big[[\mu^F+\rho\,\bar y^F_i]_{+}\big]$, an averaged form of complementary slackness that no longer forces every row to be feasible. We use the scalar parameterization as an approximation of the per-row method. The identity of \cref{prop:mass} in terms of the effective multipliers holds for either parameterization; its version in terms of $\mu$ requires per-row multipliers. In the scalar parameterization the two rollout rows share $\mu_{\mathrm{K}}$, and the dual target uses $\bar y^{\mathrm K}_i=\tfrac12(y^{\mathrm{KQ}}_i+y^{\mathrm{KV}}_i)$. Even at the scalar dual fixed point, therefore, $\mathbb{E}_i[\mu^{\mathrm{KQ,eff}}_i+\mu^{\mathrm{KV,eff}}_i]=2\mu_{\mathrm{K}}+J_{\mathrm K}$ with
  \begin{equation*}
    J_{\mathrm K}=\mathbb{E}_i\big[[\mu_{\mathrm{K}}+\rho y^{\mathrm{KQ}}_i]_{+}+[\mu_{\mathrm{K}}+\rho y^{\mathrm{KV}}_i]_{+}-2[\mu_{\mathrm{K}}+\rho\bar y^{\mathrm K}_i]_{+}\big]\ge0
  \end{equation*}
  by convexity of $[\cdot]_{+}$, with equality exactly when the two arguments have the same sign on every row. The scalar form of \cref{eq:mass}, $\mu_{\mathrm{B}}\,\mathbb{E}_i[1-\gamma m_i]+2\mu_{\mathrm{K}}=\omega_{Q}+\omega_{V}$, holds in our implementation only up to $J_{\mathrm K}$ and a covariance term between $m_i$ and $\mu^{\mathrm{B,eff}}_i$. The form with $\mu^{\mathrm{eff}}$ is exact. \Cref{app:mass} measures both: at 1M the $\mu^{\mathrm{eff}}$ form is between 0.148 and 0.154 and the scalar form between 0.115 and 0.145 against the target 0.15.
\end{remark}

\subsection{Agreement with ALBUM}
\begin{corollary}[Agreement with ALBUM]
  \label{cor:album}
  Let LBLP be a deterministic tabular instance as in \cref{prop:fixed}. If $\lambda_{\mathrm{B}}>\max_{(s,a)}\lambda^*_{\mathrm B}(s,a)$ and $\lambda_{\mathrm{E}}>\max_{(s,a)}\lambda^*_{\mathrm E}(s,a)$ for the optimal one-step multipliers $\lambda^*$ of $\Lambda$ (\cref{sec:alm}), then the ALBUM minimizer of \cref{eq:album} and every AL-LBLP fixed point coincide, for every $\lambda_{\mathrm{K}}\ge0$, $K$, and $\rho>0$.
\end{corollary}

\begin{proof}
Under the stated condition on $\lambda_{\mathrm{B}}$ and $\lambda_{\mathrm{E}}$, Proposition~4.7 of \citet{yang2026lifted} shows that the hinge relaxation on $\mathcal{M}_\mathcal{D}$ has the unique minimizer $(Q^*_\mathcal{D},V^*_\mathcal{D})$, for every $\lambda_{\mathrm{K}}\ge0$ and every rollout length $K$. Their equation~4 is proportional to \eqref{eq:album} in the tabular case, where the stop-gradient leaves the function values unchanged. By \cref{prop:fixed}, in the tabular case the primal part of every AL-LBLP fixed point is $(Q^*_\mathcal{D},V^*_\mathcal{D})$ as well, for every $\rho>0$. The two points coincide.
\end{proof}

The corollary gives a condition under which the two methods agree. Whenever the fixed coefficients of ALBUM are large enough to be exact, every AL-LBLP fixed point is the same point, and no threshold has to be known. Below the thresholds, Proposition~4.7 of \citet{yang2026lifted} no longer guarantees exactness, so the two methods may differ, while every AL-LBLP fixed point is still the LBLP solution.

\subsection{Mass identity}
\begin{proposition}[Mass identity]
  \label{prop:mass}
  Let $Q_\theta$ and $V_\phi$ each have an additive output bias, let $m_i\in\{0,1\}$ be the terminal mask of the one-step row, and let the $K$-step target be detached. At every stationary point of $\mathcal L_\rho$ in $(\theta,\phi)$, for any multipliers $\mu\ge0$, the effective multipliers $\mu^{F,\mathrm{eff}}_i=[\mu^F_i+\rho y^F_i]_{+}$ satisfy \cref{eq:mass},
  \begin{equation*}
    \mathbb{E}_i\big[(1-\gamma m_i)\,\mu^{\mathrm{B,eff}}_i+\mu^{\mathrm{KQ,eff}}_i+\mu^{\mathrm{KV,eff}}_i\big]=\omega_{Q}+\omega_{V} .
  \end{equation*}
  At a dual fixed point of \cref{eq:dual} with one multiplier per row (in particular separate multipliers for the KQ and KV rows) the same holds with $\mu$ in place of $\mu^{\mathrm{eff}}$. For constant row weights and $m\equiv1$ the left-hand side is $\lambda_{\mathrm{B}}(1-\gamma)+2\lambda_{\mathrm{K}}$, so the bounded-update condition $\lambda_{\mathrm{B}}(1-\gamma)+2\lambda_{\mathrm{K}}\ge\omega_{Q}+\omega_{V}$ of \citet[Prop.~5.1]{yang2026lifted} holds with equality for the weights that AL-LBLP applies.
\end{proposition}

\begin{proof}
Write $Q_\theta(s,a)=\tilde Q_\theta(s,a)+b_Q$ and $V_\phi(s)=\tilde V_\phi(s)+b_V$, where $b_Q$ and $b_V$ are the output biases, so that $\partial Q_\theta(s,a)/\partial b_Q=1$ and $\partial V_\phi(s)/\partial b_V=1$ at every input. In the detached program the rows are
\begin{equation*}
  y^{\mathrm B}_i=r_i+\gamma m_i V_\phi(s'_i)-Q_\theta(s_i,a_i),\quad
  y^{\mathrm{KQ}}_i=\operatorname{sg}[\hat y_{K,i}]-Q_\theta(s_i,a_i),\quad
  y^{\mathrm{KV}}_i=\operatorname{sg}[\hat y_{K,i}]-V_\phi(s_i),\quad
  y^{\mathrm E}_i=Q_\theta(s_i,a_i)-V_\phi(s_i),
\end{equation*}
where the stop-gradient makes $\hat y_{K,i}$ a constant with respect to $(\theta,\phi)$. Their partial derivatives with respect to the two biases are
\begin{equation*}
  \frac{\partial y^{\mathrm B}_i}{\partial b_Q}=-1,\;\;
  \frac{\partial y^{\mathrm{KQ}}_i}{\partial b_Q}=-1,\;\;
  \frac{\partial y^{\mathrm{KV}}_i}{\partial b_Q}=0,\;\;
  \frac{\partial y^{\mathrm E}_i}{\partial b_Q}=1;\qquad
  \frac{\partial y^{\mathrm B}_i}{\partial b_V}=\gamma m_i,\;\;
  \frac{\partial y^{\mathrm{KQ}}_i}{\partial b_V}=0,\;\;
  \frac{\partial y^{\mathrm{KV}}_i}{\partial b_V}=-1,\;\;
  \frac{\partial y^{\mathrm E}_i}{\partial b_V}=-1 .
\end{equation*}
By the chain rule and (F1), with $\mu^{F,\mathrm{eff}}_i=[\mu^F_i+\rho y^F_i]_{+}$,
\begin{align*}
  \frac{\partial\mathcal L_\rho}{\partial b_Q}&=\omega_{Q}+\mathbb{E}_i\big[-\mu^{\mathrm{B,eff}}_i-\mu^{\mathrm{KQ,eff}}_i+\mu^{\mathrm{E,eff}}_i\big],\\
  \frac{\partial\mathcal L_\rho}{\partial b_V}&=\omega_{V}+\mathbb{E}_i\big[\gamma m_i\,\mu^{\mathrm{B,eff}}_i-\mu^{\mathrm{KV,eff}}_i-\mu^{\mathrm{E,eff}}_i\big],
\end{align*}
where the $\omega$ terms come from the objective $\mathbb{E}_\mathcal{D}[\omega_{Q} Q_\theta+\omega_{V} V_\phi]$. Adding the two,
\begin{equation*}
  \frac{\partial\mathcal L_\rho}{\partial b_Q}+\frac{\partial\mathcal L_\rho}{\partial b_V}
  =\omega_{Q}+\omega_{V}-\mathbb{E}_i\big[(1-\gamma m_i)\,\mu^{\mathrm{B,eff}}_i+\mu^{\mathrm{KQ,eff}}_i+\mu^{\mathrm{KV,eff}}_i\big].
\end{equation*}
The epigraph terms cancel because $y^{\mathrm E}_i$ is invariant to a common shift of $Q$ and $V$. At a stationary point of $\mathcal L_\rho$ in $(\theta,\phi)$ both partial derivatives vanish, so the right-hand side is zero, which is \cref{eq:mass}. This holds for any $\mu\ge0$, since the multipliers enter the loss only through $\mu^{\mathrm{eff}}$.

At a dual fixed point of \cref{eq:dual}, \cref{eq:mueff_eq_mu} gives $\mu^{\mathrm{eff}}=\mu$ row by row, so \cref{eq:mass} holds with $\mu$. This step uses one multiplier per row. Suppose instead that the KQ and KV rows of window $i$ share a multiplier $\mu^{\mathrm K}_i$ updated with $\bar y^{\mathrm K}_i=\tfrac12(y^{\mathrm{KQ}}_i+y^{\mathrm{KV}}_i)$, as the scalar implementation of \cref{sec:alm} does. Then the fixed point gives $[\mu^{\mathrm K}_i+\rho\bar y^{\mathrm K}_i]_{+}=\mu^{\mathrm K}_i$, and convexity of $[\cdot]_{+}$ yields
\begin{equation*}
  \mu^{\mathrm{KQ,eff}}_i+\mu^{\mathrm{KV,eff}}_i=[\mu^{\mathrm K}_i+\rho y^{\mathrm{KQ}}_i]_{+}+[\mu^{\mathrm K}_i+\rho y^{\mathrm{KV}}_i]_{+}\ge2\mu^{\mathrm K}_i,
\end{equation*}
with equality exactly when the two arguments have the same sign. The form of \cref{eq:mass} with $\mu$ then holds only up to this nonnegative gap, which is $J_{\mathrm K}$ of \cref{rem:scalar}.

Finally, suppose every row of a family carries the same constant weight ($\mu^{\mathrm{B,eff}}_i\equiv\lambda_{\mathrm{B}}$, $\mu^{\mathrm{KQ,eff}}_i\equiv\mu^{\mathrm{KV,eff}}_i\equiv\lambda_{\mathrm{K}}$) and $m_i\equiv1$. Then the left-hand side of \cref{eq:mass} is $\lambda_{\mathrm{B}}(1-\gamma)+2\lambda_{\mathrm{K}}$, the quantity in Proposition~5.1 of \citet{yang2026lifted}. There the update is bounded along the common-shift direction only if this quantity is at least $\omega_{Q}+\omega_{V}$, and \cref{eq:mass} states that the effective weights of AL-LBLP meet it with equality at every stationary point. The identity is the same computation carried out for ALBUM's fixed coefficients by \citet[Remark~I.6]{yang2026lifted}. Without the stop-gradient on $\hat y_{K,i}$ the rollout rows depend on $b_V$ through $\gamma^K M_i V_\phi(s_{K,i})$, with $M_i$ the window's bootstrap mask, and the coefficient of $\mu^{\mathrm{KQ,eff}}_i+\mu^{\mathrm{KV,eff}}_i$ in \cref{eq:mass} becomes $1-\gamma^K M_i$.
\end{proof}

\paragraph{Reading of the identity.}
\Cref{eq:mass} is the balance of the output-bias gradients of $Q_\theta$ and $V_\phi$. The objective pushes all values down with total mass $\omega_{Q}+\omega_{V}$, and the penalty rows push back with the mass carried by the effective multipliers. In ALBUM the weight of a row is at most its fixed $\lambda$, so the penalty rows can push back with mass at most $\lambda_{\mathrm{B}}(1-\gamma)+2\lambda_{\mathrm{K}}$ for $m\equiv1$. A balance exists only if this bound reaches $\omega_{Q}+\omega_{V}$, which is the condition of Proposition~5.1 of \citet{yang2026lifted}; with ALBUM's coefficients and $\gamma=0.995$ the bound is about 2.0 against 0.15. In AL-LBLP the weights $[\mu+\rho y]_{+}$ adjust through $y$ at every step and through $\mu$ at the dual step, so the balance is reached by the optimization itself, with no condition on any coefficient. We log two residuals of \cref{eq:mass} on every training batch. The one with $\mu^{\mathrm{eff}}$ checks the identity directly at the current parameters. The one with the scalar $\mu$ in its place, $\mu_{\mathrm{B}}\,\mathbb{E}_i[1-\gamma m_i]+2\mu_{\mathrm{K}}-(\omega_{Q}+\omega_{V})$, measures how far the dual is from its fixed point, but in the scalar parameterization it is biased downward by the gap $J_{\mathrm K}$ of \cref{rem:scalar}. \Cref{app:mass} reports both at 1M steps for every environment of \cref{tab:main}.

\subsection{Invariance of the advantage-normalized actor}
We write the actor loss \eqref{eq:actor} with the denominator $D_A=\operatorname{sg}\,\mathbb{E}_s|Q(s,\pi(s))-V(s)|$ and compare it with the TD3+BC loss, whose denominator is $D_Q=\operatorname{sg}\,\mathbb{E}_s|Q(s,\pi(s))|$.

\begin{assumption}
  \label{ass:continuing}
  The task is continuing: timeouts are not terminals, and every transition has mask $m=1$ (and every window has bootstrap mask $M=1$). The pair $(Q,V)$ ranges over a class of functions that contains $(cQ+k,\,cV+k)$ for every $c>0$ and $k\in\mathbb{R}$ whenever it contains $(Q,V)$, for example networks with output biases. Every bootstrap value $V(s')$ and $V(s_K)$ in the rows of \cref{eq:lblp} is the value of $V$ itself, and no state has a fixed value.
\end{assumption}

\begin{proposition}[Invariance to affine reward transformations]
  \label{prop:invariance}
  Under \cref{ass:continuing}, let $r'=c\,r+b$ with $c>0$ and set $k=b/(1-\gamma)$.
  \begin{enumerate}
    \item[(i)] If $(Q,V)$ is an optimal solution of LBLP \eqref{eq:lblp} over this class for the reward $r$, then $(cQ+k,\,cV+k)$ is an optimal solution for $r'$.
    \item[(ii)] The $|A|$-normalized actor update of \cref{eq:actor} is the same at $(Q,V)$ and at $(cQ+k,\,cV+k)$, hence, by (i), the same at the solutions for $r$ and $r'$.
    \item[(iii)] The $|Q|$-normalized update gives the same gradient for every distribution of $Q$ only if $b=0$, and it is unchanged when $b=0$, since then $D_{Q'}=c\,D_Q$.
  \end{enumerate}
  Moreover, adding the same action-independent shift $\delta(s)$ to $Q$ and $V$ leaves the $|A|$-normalized update unchanged, while it rescales the $Q$ term of the $|Q|$-normalized update by $\mathbb{E}_s|Q(s,\pi(s))|/\mathbb{E}_s|Q(s,\pi(s))+\delta(s)|$.
\end{proposition}

\begin{proof}
(i) Put $(Q',V')=(cQ+k,\,cV+k)$ and $r'=c\,r+b$ into the rows of \cref{eq:lblp}. By \cref{ass:continuing} all masks equal one and every bootstrap value is the value of $V$ itself, so it becomes $cV+k$. With $\Gamma=\gamma^K$ and $R=\sum_{t<K}\gamma^t r_t$, the shifted $K$-step target is $\hat y'_K=cR+b\sum_{t<K}\gamma^t+\Gamma(cV(s_K)+k)=c\hat y_K+(1-\Gamma)k+\Gamma k=c\hat y_K+k$, because $b\sum_{t<K}\gamma^t=b(1-\Gamma)/(1-\gamma)=(1-\Gamma)k$. Hence
\begin{align*}
  y'^{\mathrm B}&=(c\,r+b)+\gamma(cV(s')+k)-(cQ+k)=c\,y^{\mathrm B}+\big[b-(1-\gamma)k\big]=c\,y^{\mathrm B},\\
  y'^{\mathrm{KQ}}&=(c\hat y_K+k)-(cQ+k)=c\,y^{\mathrm{KQ}},\qquad
  y'^{\mathrm{KV}}=(c\hat y_K+k)-(cV+k)=c\,y^{\mathrm{KV}},\\
  y'^{\mathrm E}&=(cQ+k)-(cV+k)=c\,y^{\mathrm E}.
\end{align*}
Since $c>0$, every row is violated under $r'$ at $(Q',V')$ if and only if it is violated under $r$ at $(Q,V)$. By \cref{ass:continuing} the class contains $(Q',V')$ and is closed under the inverse map $(Q',V')\mapsto((Q'-k)/c,\,(V'-k)/c)$, which has the same form. The map $(Q,V)\mapsto(cQ+k,cV+k)$ is therefore a bijection between the two feasible sets. The objective is linear with positive weights, and $\mathbb{E}_\mathcal{D}[\omega_{Q}Q'+\omega_{V}V']=c\,\mathbb{E}_\mathcal{D}[\omega_{Q}Q+\omega_{V}V]+(\omega_{Q}+\omega_{V})k$, so $c>0$ preserves the order of objective values. Minimizers correspond under the map, which is (i).

(ii) The advantage transforms as $A'=Q'-V'=cA$, so $D_{A'}=c\,D_A$ and $A'(s,\pi(s))/D_{A'}=A(s,\pi(s))/D_A$: the $Q$ term of \cref{eq:actor} is unchanged. If the numerator is written with $Q$, as in our implementation, then $Q'(s,\pi(s))/D_{A'}=Q(s,\pi(s))/D_A+k/(c\,D_A)$, and the second term is a constant with respect to the policy parameters because $D_A$ is under stop-gradient. The behavior-cloning term does not involve the reward. The gradients coincide.

(iii) The denominator $D_{Q'}=\mathbb{E}_s|cQ(s,\pi(s))+k|$ depends on $k$. If $Q$ has one sign on the batch and $k$ has the same sign, then $D_{Q'}=c\,D_Q+|k|$, and the $Q$ term of the update is scaled by $c\,D_Q/(c\,D_Q+|k|)\ne1$ whenever $k\ne0$. The update is therefore unchanged for every $Q$ only if $b=0$. Conversely, if $b=0$ then $k=0$, $Q'=cQ$, and $D_{Q'}=c\,D_Q$, so $Q'(s,\pi(s))/D_{Q'}=Q(s,\pi(s))/D_Q$ and the update is unchanged.

For the shift, $Q'(s,a)=Q(s,a)+\delta(s)$ and $V'(s)=V(s)+\delta(s)$ give $A'=A$, so $D_{A'}=D_A$, and $Q'(s,\pi(s))=Q(s,\pi(s))+\delta(s)$ differs from $Q(s,\pi(s))$ by a term that is constant in the policy parameters. The $|Q|$ denominator changes from $\mathbb{E}_s|Q(s,\pi(s))|$ to $\mathbb{E}_s|Q(s,\pi(s))+\delta(s)|$, which rescales the $Q$ term by the stated ratio.
\end{proof}

\Cref{prop:invariance} is a statement about invariance, not about performance. With true terminals a reward shift changes the optimal policy, so shift invariance is neither expected nor desired there, which is why \cref{ass:continuing} is needed. For the same reason part (i) does not cover the tabular program on $\mathcal{M}_\mathcal{D}$ of \cref{sec:prelim}: a successor state that the data never leave has the fixed value zero there and acts as a terminal. The effect of the normalization on the same critic is measured in \cref{sec:norm}.

\newpage
\section{Experimental details}
\label{app:details}
Each subsection lists only what differs from the protocol of \cref{sec:experiments}. The fixed hyperparameters are in \cref{tab:hparams} and the per-environment selections in \cref{tab:side-picks,tab:alpha-visual}.

\begin{table*}[h]
  \centering
  \caption{Hyperparameters of AL-LBLP (last block: the pixel-based tasks of \cref{tab:visual}); $\rho$ and $\alpha$ per environment in \cref{tab:side-picks,tab:alpha-visual}.}
  \label{tab:hparams}
    \begin{small}
      \begin{tabular}{lp{0.62\textwidth}}
        \toprule
        Hyperparameter & Value \\
        \midrule
        Rollout length $K$ & 10 \\
        Discount $\gamma$ & 0.995 \\
        Penalty parameter $\rho$ & per environment; (\cref{tab:side-picks}, \cref{tab:alpha-visual}) \\
        Objective weights $(\omega_{Q},\omega_{V})$ & $(0.1,\,0.05)$ \\
        Multiplier parameterization $(\mu_B, \mu_E, \mu_K)$ & initialized at $10^{-3}$ \\
        Optimizer & Adam \\
        Learning rate & $3\times10^{-4}$ \\
        Networks & MLP; $4\times512$ \\
        LayerNorm & only $Q$ and $V$ \\
        Critic ensemble / target network & none / none \\
        Batch size & 256 \\
        Gradient steps / Evaluation & 1M / 50 episodes \\
        Actor objective & \cref{eq:actor}, advantage normalization in every environment \\
        \midrule
        \multicolumn{2}{l}{\emph{Pixel-based tasks}} \\
        \midrule
        Encoder & small IMPALA \\
        Discount $\gamma$ & 0.99 \\
        Critic optimizer & Adam with $\beta_{2}=0.99$; gradient-norm clip 10 \\
        Gradient steps / Evaluation & 500k / 50 episodes \\
        \bottomrule
      \end{tabular}
    \end{small}
\end{table*}

\subsection{State-based tasks}
We use the \textit{singletask} datasets of the ten OGBench environments and report baselines and our method on the five tasks of each environment (\cref{sec:main}). The offline-to-online (\cref{sec:off2on}) and ablation (\cref{sec:ablations}) experiments use the default task of each environment. We train a single $Q$ and a single $V$ without a target network for 1M gradient steps with batch 256, $\gamma=0.995$, $K=10$. We report the evaluation at 1M with 50 episodes, over 8 seeds. The selection rule is the highest mean success at 1M on the default task over 3 seeds, with ties going to the smaller $\rho$ and then to the smaller value of $\alpha$. The Gaussian policy selects $(\rho,\alpha)$ from $\rho\in\{0.05,0.1,0.5\}\times\alpha\in\{1,3,10\}$ (nine combinations). The flow policy selects $(\rho,\alpha_{\mathrm{flow}})$ from $\rho\in\{0.1,0.5\}\times\alpha_{\mathrm{flow}}\in\{3,10,30\}$ (six). Best-of-$N$ selects $(\rho,N)$ from $\rho\in\{0.1,0.5\}\times N\in\{8,16,32\}$ (six) and acts on the checkpoints of flow-policy runs with its $\rho$. The nine combinations were fixed by a budget analysis on the same default-task runs, so the grid itself is chosen in-sample. The selected values are in \cref{tab:side-picks}, and the detailed hyperparameters of AL-LBLP are listed in \cref{tab:hparams}.

\begin{table}[h]
  \centering
  \caption{Per-environment configuration of the three AL-LBLP columns of \cref{tab:main} (grids in \cref{app:details}), selected on the default task by the highest 3-seed mean at 1M; ties to the smaller $\rho$, then the smaller value.}
  \label{tab:side-picks}
    \begin{small}
      \resizebox{\columnwidth}{!}{%
      \begin{tabular}{llcccccccccc}
        \toprule
        & & antmaze-l & antmaze-g & humanoid-m & humanoid-l & antsoccer & cube-s & cube-d & scene & puzzle-3 & puzzle-4 \\
        \midrule
        Gaussian policy & $\rho$ & 0.05 & 0.1 & 0.1 & 0.5 & 0.5 & 0.05 & 0.5 & 0.5 & 0.05 & 0.05 \\
         & $\alpha$ & 3 & 1 & 1 & 3 & 3 & 10 & 10 & 3 & 3 & 10 \\
        \midrule
        Flow policy & $\rho$ & 0.1 & 0.1 & 0.1 & 0.5 & 0.5 & 0.1 & 0.5 & 0.5 & 0.1 & 0.1 \\
         & $\alpha_{\mathrm{flow}}$ & 3 & 3 & 10 & 30 & 10 & 30 & 30 & 10 & 30 & 30 \\
        \midrule
        Best-of-$N$ & $\rho$ & 0.5 & 0.1 & 0.1 & 0.1 & 0.5 & 0.5 & 0.5 & 0.5 & 0.1 & 0.5 \\
         & $N$ & 8 & 8 & 32 & 16 & 16 & 8 & 8 & 8 & 16 & 16 \\
        \bottomrule
      \end{tabular}
      }
    \end{small}
\end{table}

\subsection{Baselines}
IQL, ReBRAC, FQL, and IFQL are taken from \citet{park2025flow}, who use $\gamma=0.99$ on six of the ten environments and report the mean of the last three evaluations. FAC \citep{chae2026flow} reports one evaluation after 1M steps with 25 episodes. DROL \citep{mu2026preserve} states neither its number of seeds nor its evaluation rule. Our evaluation protocol follows ALBUM \citep{yang2026lifted}, which reports one evaluation after 1M steps with 50 episodes over 8 seeds, and we take the ALBUM results from its paper. We do not include floq \citep{agrawalla2025floq}, which reports 2M gradient steps with 3 seeds.

\subsection{Pixel-based tasks}
We run task 1 of the five pixel-based manipulation environments following the pixel configuration of FQL: a small IMPALA encoder, three stacked $64\times64$ frames, and random-crop augmentation (last block of \cref{tab:hparams}). Training runs for 500k gradient steps with $\gamma=0.99$, a critic Adam $\beta_2$ of 0.99, and gradient-norm clipping at 10, over 4 seeds (seeds 0--3). We use the Gaussian policy for pixel-based tasks and $(\rho,\alpha)$ is selected per environment from $\rho\in\{0.5,1,2\}\times\alpha\in\{10,30,100\}$ by the mean final evaluation over the same four seeds that are reported (\cref{tab:alpha-visual}). The baselines IQL, ReBRAC, and FQL are from \citet{park2025flow} and FAC from \citet{chae2026flow}. The results are in \cref{tab:visual} and the sensitivity to $\rho$ and $\alpha$ is in \cref{app:visual}.

\begin{table}[h]
  \centering
  \caption{$(\rho,\alpha)$ of \cref{tab:visual}, selected per environment from $\{0.5,1,2\}\times\{10,30,100\}$ by the mean final evaluation over the reported seeds; ties to the smaller $\rho$, then $\alpha$.}
  \label{tab:alpha-visual}
    \begin{small}
      \begin{tabular}{lccccc}
        \toprule
        & cube-single & cube-double & scene & puzzle-3x3 & puzzle-4x4 \\
        \midrule
        $\rho$ & 1 & 2 & 1 & 0.5 & 0.5 \\
        $\alpha$ & 100 & 30 & 10 & 10 & 10 \\
        \bottomrule
      \end{tabular}
    \end{small}
\end{table}

\subsection{Offline-to-online fine-tuning}
We follow the offline-to-online protocol of FQL \citep{park2025flow}. Each run restores the 1M offline checkpoint of the default task from \cref{tab:main} and trains for 1M online steps with one gradient step per environment step. Online transitions are appended to the dataset and sampled uniformly. Each policy keeps its offline hyperparameters (\cref{tab:side-picks}) and tuned hyperparameters (\textit{i.e.,} $\rho$, $\alpha$, $\alpha_{\mathrm{flow}}$ and $N$), over 3 seeds. The Gaussian policy explores with $\mathrm{clip}(0.2\,\epsilon,-0.5,0.5)$, $\epsilon\sim\mathcal N(0,I)$, as in ReBRAC. The flow policy acts without added noise, and best-of-$N$ also executes the largest-$Q$ of its $N$ samples online. Evaluation uses 50 episodes at every 100k online steps (\cref{fig:combined_results}(a)).

\subsection{Long-horizon navigation tasks and large batch size}
For long-horizon tasks, \textit{antmaze-giant} and \textit{humanoidmaze-giant}, we increase $\gamma=0.995$ and $K=10$ to $\gamma=0.999$ and $K=50$, the only deviation from the shared protocol. $(\rho,\alpha)$ is selected on task 1 with 3 seeds from the same nine combinations as in the shared protocol, $\rho\in\{0.05,0.1,0.5\}\times\alpha\in\{1,3,10\}$, so the tuning budget is unchanged. This gives $(0.05,1)$ on \textit{humanoidmaze-giant} (63.3 at 1M) and $(0.1,1)$ on \textit{antmaze-giant} (40.0 at 1M, against 35.3 for $(0.05,1)$). On \textit{humanoidmaze-giant}, changing only one of $K$ and $\gamma$ at $\rho=0.5$ leaves the task unsolved at 1M (task 1, 3 seeds): 0.7 with $K=50$ at $\gamma=0.995$ and 0.0 with $\gamma=0.999$ at $K=10$. The ALBUM control uses its original configuration (fixed coefficients, critic learning rate $10^{-4}$, single $Q$, batch 256) with only $K=50$ and $\gamma=0.999$ changed, and sweeps $\alpha\in\{0.003,0.01,0.03,0.1,0.3,1\}$ on task 1 with 3 seeds; all 36 runs end at 0.0 success with a mean critic value between $-999.3$ and $-999.0$, close to $-1/(1-\gamma)=-1000$. Since all values tie, we use the smallest, $\alpha=0.003$, and train all five tasks of both environments with 8 seeds at batch 256.

In \cref{tab:longhorizon}, \cref{tab:batch-fql} and \cref{tab:longhorizon-tasks}, we train AL-LBLP for all five tasks with 8 seeds over 1M steps at batch sizes 256, 2048, 4096, and 8192, with $(\rho, \alpha)$ of batch 256 (not re-tuned), and evaluate once at 1M. The $^\dagger$ rows of \cref{tab:longhorizon} instead re-select $(\rho,\alpha)$ at each batch size within the same budget: $(0.1,1)$ at batch 256 and $(0.5,1)$ at 2048, 4096, and 8192 on \textit{antmaze-giant}, and $(0.05,1)$ at batch 256 and $(0.1,1)$ at 2048, 4096, and 8192 on \textit{humanoidmaze-giant}. The $K=10$ reference rows of \cref{tab:longhorizon} use the shared protocol ($\gamma=0.995$, $K=10$) with 8 seeds. \textit{antmaze-giant} uses $(\rho,\alpha)=(0.1,1)$ and \textit{humanoidmaze-large} uses $(0.5,3)$.

The AL-LBLP rows of \cref{tab:batch-fql} are these $K=10$ runs. The FQL rows of \cref{tab:batch-fql} run the authors' code with its README settings and only the batch size changed, over 3 seeds; the batch-256 values are the published ones. The FAC rows run the authors' code in the same way, with the BC proxy trained as at batch 256, over 3 seeds; their batch-256 values are also the published ones.

\begin{table}[h]
  \centering
  \caption{Changing only the batch size under the shared protocol ($\gamma=0.995$, $K=10$): five-task success (\%) at 1M, mean $\pm$ std. AL-LBLP: 8 seeds. FQL and FAC: 3 seeds; batch 256: published (8 seeds). --: not run. Bold: within 95\% of each environment's best.}
  \label{tab:batch-fql}
    \begin{small}
        \begin{tabular}{llcccc}
          \toprule
          & & \multicolumn{4}{c}{Batch size} \\
          \cmidrule(lr){3-6}
          Environment & Method & 256 & 2048 & 4096 & 8192 \\
          \midrule
          humanoidmaze-large & FQL & 4 $\pm$2 & 4.9 $\pm$6.3 & 0.0 $\pm$0.0 & -- \\
           & FAC & 8.3 $\pm$5.5 & 0.0 $\pm$0.0 & 0.0 $\pm$0.0 & -- \\
           & AL-LBLP & 17.4 $\pm$1.0 & 32.9 $\pm$7.0 & \textbf{39.9} $\pm$6.5 & 33.1 $\pm$7.5 \\
          \midrule
          antmaze-giant & FQL & 9 $\pm$6 & 0.0 $\pm$0.0 & 0.0 $\pm$0.0 & -- \\
           & FAC & 23.0 $\pm$5.1 & 0.1 $\pm$0.2 & 0.0 $\pm$0.0 & -- \\
           & AL-LBLP & 47.5 $\pm$7.4 & \textbf{56.1} $\pm$4.0 & \textbf{53.7} $\pm$3.9 & 52.2 $\pm$4.5 \\
          \bottomrule
        \end{tabular}
    \end{small}
\end{table}

\subsection{Ablations}
All ablations use the default task of each of the ten environments with 3 seeds. For the penalty parameter we evaluate $\rho\in\{0.01,0.05,0.1,0.5,1\}$ at 1M (\cref{fig:rho-norm}(a)) and train $\rho\in\{0.1,0.5,1\}$ to 3M, evaluating every 0.5M (\cref{fig:combined_results}(b), \cref{tab:rho-speed}). For each $(\rho,\text{environment})$ we use the $\alpha$ selected at 1M over $\{1,3,10,30,100\}$. For the actor normalization (\cref{fig:rho-norm}(b)--(d)) we fix $\rho=0.1$ and attach three actors to the same critic: $|A|$ with $\alpha\in\{1,3,10,30,100\}$, $|Q|$ with $\alpha\in\{0.003,0.01,0.03,0.1,0.3,1\}$, and no normalization with $\alpha\in\{1,3,10,30,100\}$. Because the critic does not read the actor, the logged $Q$, $V$, and multiplier values of the three actors are identical for every environment and seed.

The violation statistics of \cref{tab:rho-viol} are measured on the training batches of the 100k steps preceding 1M; $\mathbb{E}[y\mid y>0]$ is the ratio of the two statistics computed per seed and then averaged, and the $K$-step family is the mean of the KQ and KV rows. The ALBUM column (reference) uses ALBUM's fixed coefficients $\{\lambda_B: 2.5, \lambda_E: 2.5, \lambda_K: 1\}$, its own critic learning rate ($0.0001$), and the $|Q|$ actor normalization. The row weight of \cref{tab:rho-speed} is the batch maximum over the batch mean of the effective weight $[\mu+\rho y]_{+}$, averaged over the log rows in $(0,100\mathrm{k}]$. The srank is the number of singular values of the critic's 512 penultimate features (20{,}000 samples) that make up 99\% of their sum \citep{kumar2021implicit}, measured at 0.5M. The statistics of \cref{tab:goal-rows} are measured instead at the 1M checkpoints (five tasks, 8 seeds), on 512{,}000 windows per task sampled uniformly from the dataset, with the KQ row only. A goal window contains the transition into the goal, so its target $\hat y_K$ has no bootstrap term (mean $-13.64$ on \textit{antmaze-giant} and $-17.34$ on \textit{humanoidmaze-giant}). The ALBUM rows are the runs of \cref{sec:longhorizon} with $\alpha=0.003$.

\begin{table}[t]
  \centering
  \caption{Violations at 1M on the 10 default tasks, mean $\pm$ std over 3 seeds: fraction $P(y>0)$ and size $\mathbb{E}[y\mid y>0]$ of violated rows in training batches. Bold: lowest (within 5\%).}
  \label{tab:rho-viol}
    \begin{small}
        \begin{tabular}{llcccc}
          \toprule
          & & \multicolumn{3}{c}{AL-LBLP} & ALBUM \\
          \cmidrule(lr){3-5}\cmidrule(lr){6-6}
          Family & & $\rho=0.1$ & $\rho=0.5$ & $\rho=1$ & \\
          \midrule
          \multirow{2}{*}{Bellman} & $P(y>0)$ (\%) & 47.3 $\pm$0.9 & 39.9 $\pm$1.0 & 37.3 $\pm$0.4 & \textbf{33.0} $\pm$0.9 \\
           & $\mathbb{E}[y\mid y>0]$ & 1.38 $\pm$0.06 & 0.82 $\pm$0.04 & \textbf{0.63} $\pm$0.03 & 0.84 $\pm$0.15 \\
          \midrule
          \multirow{2}{*}{Epigraph} & $P(y>0)$ (\%) & 46.8 $\pm$0.9 & 42.3 $\pm$0.7 & 37.1 $\pm$0.5 & \textbf{31.7} $\pm$2.8 \\
           & $\mathbb{E}[y\mid y>0]$ & 1.41 $\pm$0.07 & 0.80 $\pm$0.02 & \textbf{0.60} $\pm$0.02 & \textbf{0.59} $\pm$0.07 \\
          \midrule
          \multirow{2}{*}{$K$-step} & $P(y>0)$ (\%) & 34.7 $\pm$0.6 & 15.2 $\pm$0.7 & 9.6 $\pm$0.1 & \textbf{7.0} $\pm$1.8 \\
           & $\mathbb{E}[y\mid y>0]$ & 1.18 $\pm$0.03 & 0.80 $\pm$0.04 & \textbf{0.71} $\pm$0.02 & 5.17 $\pm$2.53 \\
          \bottomrule
        \end{tabular}
    \end{small}
\end{table}

The path ablation of \cref{tab:album-path} uses $\rho=1$ and a critic learning rate of $3\times10^{-4}$ for the AL-LBLP rows, evaluated at 1M and 3M, with $\alpha$ re-selected per row and environment at 1M; the $|Q|$ rows use the six-value grid ($\alpha\in\{0.003,0.01,0.03,0.1,0.3,1\}$). Its third row keeps the quadratic term but fixes the multipliers at ALBUM's coefficients $\{\lambda_B: 2.5, \lambda_E: 2.5, \lambda_K: 1\}$. Its fourth keeps the learned multipliers but replaces the quadratic term by the hinge. Its fifth repeats the fourth with the critic learning rate of ALBUM ($10^{-4}$); the actor and dual learning rates stay at $3\times10^{-4}$. The ALBUM row is the original implementation of ALBUM with its own $\alpha$. The last column is the $K$-step $\mathbb{E}[y\mid y>0]$ of \cref{tab:rho-viol} at 1M: $y^{\mathrm{K}}$ are the residuals of the KQ and KV rows, with the statistics averaged over the two (not $\bar y^{\mathrm{K}}$ of \cref{sec:alm}).

\subsection{Implementation}
\label{app:impl}
The supplementary code implements AL-LBLP in JAX with the settings of \cref{tab:hparams}. It contains the critic with its multipliers (\cref{alg:allblp}), the $|A|$-normalized Gaussian actor, and the evaluation on OGBench. In the numerator of \cref{eq:actor} the code uses $Q$ in place of $A$; since $V(s)$ does not depend on $\pi$, the gradient is the same. On one H100, a run of 1M steps on \textit{antmaze-large} takes about 0.24 GPU-hours without evaluation (batch 256).

\paragraph{Training several values of $\alpha$ at once.}
Because the critic is trained only at dataset actions and never evaluates the policy, the code can also train the actors for several values of $\alpha$ with one shared critic in a single run. The same property lets the critic be trained separately from the policy that uses it, for example a VLA model that selects among its own samples by best-of-$N$ (\cref{sec:intro}). \Cref{tab:alpha-speed} compares the time to train $n_\alpha$ values of $\alpha$ at once. For $n_\alpha=6$, one run with the shared critic takes 0.33 GPU-hours, against 1.41 for separate runs (4.3$\times$).

\begin{table}[h]
  \centering
  \caption{GPU-hours to train $n_\alpha$ values of $\alpha$ for 1M steps (one H100, \textit{antmaze-large}, batch 256, evaluation excluded). The critic does not read the actor, so one run can train a shared critic with $n_\alpha$ actors.}
  \label{tab:alpha-speed}
    \begin{small}
      \setlength{\tabcolsep}{4pt}
      \begin{tabular}{@{}lccccc@{}}
        \toprule
        & Critics & \multicolumn{2}{c}{GPU-h} & \multicolumn{2}{c}{Speed-up} \\
        \cmidrule(lr){3-4}\cmidrule(lr){5-6}
        & per run & $n_\alpha=3$ & $n_\alpha=6$ & $n_\alpha=3$ & $n_\alpha=6$ \\
        \midrule
        \multicolumn{6}{@{}l}{Separate runs} \\
        \quad one at a time & 1 & 0.71 & 1.41 & 1.0$\times$ & 1.0$\times$ \\
        \quad concurrent & 1 & 0.77 & 1.55 & 0.9$\times$ & 0.9$\times$ \\
        \multicolumn{6}{@{}l}{One run} \\
        \quad a critic per $\alpha$ & $n_\alpha$ & 0.38 & 0.56 & 1.8$\times$ & 2.5$\times$ \\
        \quad shared critic (\textbf{ours}) & 1 & \textbf{0.28} & \textbf{0.33} & \textbf{2.5}$\times$ & \textbf{4.3}$\times$ \\
        \bottomrule
      \end{tabular}
    \end{small}
\end{table}

\newpage
\section{Long-horizon navigation tasks}
\label{app:longhorizon}
\Cref{tab:longhorizon-tasks} lists the tasks behind the $\gamma=0.999$ and $K=50$ rows of \cref{tab:longhorizon} and compares AL-LBLP with LQL \citep{abraham2026long} and TRL \citep{park2026transitive}. The selection of $(\rho,\alpha)$ is described in \cref{app:details}.

\begin{table*}[h]
  \centering
  \caption{The $\gamma=0.999$ and $K=50$ rows of \cref{tab:longhorizon} per task: success (\%) at 1M steps, mean $\pm$ std over 8 seeds. $^\dagger$: $(\rho,\alpha)$ re-selected at each batch size within the same budget, as in \cref{tab:longhorizon}. Bold: within 95\% of the best of our four batch sizes. ALBUM: $\alpha$ selected at batch 256 and kept at batch 4096, shown in the rows without $^\dagger$. LQL \citep{abraham2026long}: offline-to-online single-task RL, 2M steps in total, mean over 4 seeds (task-level results are reported only for \textit{humanoidmaze-giant}). TRL \citep{park2026transitive}: offline goal-conditioned RL on a 1B-transition dataset, mean $\pm$ std over 4 seeds. --: not reported.}
  \label{tab:longhorizon-tasks}
    \begin{small}
    \resizebox{\columnwidth}{!}{%
      \setlength{\tabcolsep}{4pt}
      \begin{tabular}{lcccc|cc|c|c}
        \toprule
        & \multicolumn{4}{c}{AL-LBLP (ours), offline RL} & \multicolumn{2}{c}{ALBUM, offline RL} & Off2On RL & Offline GCRL \\
        \cmidrule(lr){2-5}\cmidrule(lr){6-7}\cmidrule(lr){8-8}\cmidrule(lr){9-9}
        Task & batch 256 & batch 2048 & batch 4096 & batch 8192 & batch 256 & batch 4096 & LQL & TRL \\
        \midrule
        antmaze-giant task1 & 45.5 $\pm$9.5 & \textbf{56.8} $\pm$6.1 & 47.2 $\pm$11.9 & 37.0 $\pm$14.1 & 0.2 $\pm$0.7 & 0.0 $\pm$0.0 & -- & -- \\
        antmaze-giant task2 & \textbf{82.5} $\pm$4.6 & \textbf{80.2} $\pm$5.5 & 78.0 $\pm$3.7 & 75.5 $\pm$6.2 & 82.5 $\pm$9.7 & 87.8 $\pm$5.4 & -- & -- \\
        antmaze-giant task3 & 39.8 $\pm$12.3 & \textbf{47.2} $\pm$9.7 & 35.5 $\pm$8.5 & 25.2 $\pm$10.2 & 0.0 $\pm$0.0 & 50.0 $\pm$15.3 & -- & -- \\
        antmaze-giant task4 & 75.8 $\pm$9.5 & \textbf{83.5} $\pm$6.5 & \textbf{82.5} $\pm$5.0 & 70.5 $\pm$6.6 & 63.8 $\pm$9.5 & 78.8 $\pm$9.6 & -- & -- \\
        antmaze-giant task5 & 79.5 $\pm$12.3 & \textbf{86.2} $\pm$4.5 & \textbf{86.8} $\pm$5.2 & \textbf{89.5} $\pm$5.8 & 1.0 $\pm$1.5 & 0.5 $\pm$0.9 & -- & -- \\
        \cmidrule(lr){1-9}
        five-task mean & 64.6 $\pm$3.8 & \textbf{70.8} $\pm$3.7 & 66.0 $\pm$2.2 & 59.5 $\pm$3.6 & 29.5 $\pm$2.8 & 43.4 $\pm$2.9 & -- & -- \\
        \midrule
        antmaze-giant$^\dagger$ task1 & 45.5 $\pm$9.5 & \textbf{69.8} $\pm$8.0 & \textbf{71.8} $\pm$7.0 & 65.2 $\pm$8.8 & 0.2 $\pm$0.7 & -- & -- & -- \\
        antmaze-giant$^\dagger$ task2 & \textbf{82.5} $\pm$4.6 & \textbf{86.2} $\pm$6.6 & \textbf{86.2} $\pm$3.8 & \textbf{86.2} $\pm$5.8 & 82.5 $\pm$9.7 & -- & -- & -- \\
        antmaze-giant$^\dagger$ task3 & 39.8 $\pm$12.3 & 62.8 $\pm$7.5 & \textbf{68.8} $\pm$10.3 & 61.8 $\pm$6.1 & 0.0 $\pm$0.0 & -- & -- & -- \\
        antmaze-giant$^\dagger$ task4 & 75.8 $\pm$9.5 & \textbf{84.5} $\pm$7.1 & \textbf{80.8} $\pm$5.2 & \textbf{82.8} $\pm$5.5 & 63.8 $\pm$9.5 & -- & -- & -- \\
        antmaze-giant$^\dagger$ task5 & 79.5 $\pm$12.3 & \textbf{88.5} $\pm$5.0 & \textbf{91.5} $\pm$3.3 & \textbf{91.0} $\pm$4.4 & 1.0 $\pm$1.5 & -- & -- & -- \\
        \cmidrule(lr){1-9}
        $^\dagger$ five-task mean & 64.6 $\pm$3.8 & \textbf{78.4} $\pm$3.0 & \textbf{79.8} $\pm$2.5 & \textbf{77.4} $\pm$2.5 & 29.5 $\pm$2.8 & -- & 65.4 & -- \\
        \midrule
        humanoidmaze-giant task1 & 64.8 $\pm$5.8 & \textbf{79.8} $\pm$3.1 & \textbf{80.5} $\pm$8.7 & \textbf{80.8} $\pm$6.5 & 0.0 $\pm$0.0 & 0.0 $\pm$0.0 & -- & -- \\
        humanoidmaze-giant task2 & \textbf{83.8} $\pm$3.5 & \textbf{83.2} $\pm$29.1 & \textbf{85.2} $\pm$17.7 & \textbf{85.5} $\pm$22.2 & 5.0 $\pm$2.1 & 13.0 $\pm$8.2 & -- & -- \\
        humanoidmaze-giant task3 & 43.2 $\pm$7.1 & \textbf{50.5} $\pm$10.4 & 47.2 $\pm$15.1 & 44.8 $\pm$10.9 & 0.2 $\pm$0.7 & 13.0 $\pm$8.2 & -- & -- \\
        humanoidmaze-giant task4 & 80.0 $\pm$7.4 & 84.0 $\pm$5.8 & \textbf{90.8} $\pm$4.4 & \textbf{91.2} $\pm$2.8 & 0.0 $\pm$0.0 & 5.2 $\pm$4.3 & -- & -- \\
        humanoidmaze-giant task5 & 83.0 $\pm$7.4 & \textbf{91.8} $\pm$2.7 & \textbf{94.5} $\pm$4.6 & \textbf{94.2} $\pm$2.9 & 0.8 $\pm$1.0 & 4.8 $\pm$13.4 & -- & -- \\
        \cmidrule(lr){1-9}
        five-task mean & 71.0 $\pm$3.5 & \textbf{77.8} $\pm$7.7 & \textbf{79.7} $\pm$4.3 & \textbf{79.3} $\pm$5.7 & 1.2 $\pm$0.5 & 7.2 $\pm$1.9 & -- & -- \\
        \midrule
        humanoidmaze-giant$^\dagger$ task1 & 64.8 $\pm$5.8 & 80.5 $\pm$4.9 & 82.5 $\pm$6.2 & \textbf{87.0} $\pm$5.8 & 0.0 $\pm$0.0 & -- & 31.3 & 71 $\pm$15 \\
        humanoidmaze-giant$^\dagger$ task2 & 83.8 $\pm$3.5 & \textbf{93.8} $\pm$4.2 & \textbf{93.2} $\pm$4.3 & \textbf{90.0} $\pm$6.0 & 5.0 $\pm$2.1 & -- & 97.3 & 87 $\pm$6 \\
        humanoidmaze-giant$^\dagger$ task3 & 43.2 $\pm$7.1 & \textbf{48.8} $\pm$10.7 & \textbf{47.0} $\pm$21.9 & 42.0 $\pm$15.6 & 0.2 $\pm$0.7 & -- & 70.7 & 44 $\pm$8 \\
        humanoidmaze-giant$^\dagger$ task4 & 80.0 $\pm$7.4 & \textbf{90.8} $\pm$5.4 & \textbf{91.5} $\pm$6.2 & \textbf{92.0} $\pm$3.2 & 0.0 $\pm$0.0 & -- & 80.7 & 94 $\pm$4 \\
        humanoidmaze-giant$^\dagger$ task5 & 83.0 $\pm$7.4 & \textbf{90.0} $\pm$4.5 & \textbf{93.0} $\pm$4.4 & \textbf{92.8} $\pm$4.1 & 0.8 $\pm$1.0 & -- & 98.7 & 99 $\pm$1 \\
        \cmidrule(lr){1-9}
        $^\dagger$ five-task mean & 71.0 $\pm$3.5 & \textbf{80.8} $\pm$3.6 & \textbf{81.5} $\pm$5.9 & \textbf{80.8} $\pm$3.2 & 1.2 $\pm$0.5 & -- & 75.7 & 79 $\pm$2 \\
        \bottomrule
      \end{tabular}
    }
    \end{small}
\end{table*}

\subsection{Comparison with LQL}
LQL \citep{abraham2026long} reports 65.4 on \textit{antmaze-giant}, where its best actor, a Gaussian policy, trains online for 2M steps with the offline data as prior data, and 75.7 on \textit{humanoidmaze-giant}, where it trains 1M steps offline and 1M online on a 100M-transition dataset; both are the success at the end of online training over 4 seeds. AL-LBLP trains only for 1M offline gradient steps, without environment interaction, on the standard datasets (1M transitions for \textit{antmaze-giant} and 4M for \textit{humanoidmaze-giant}), and at its best batch size reaches 79.8 (batch 4096) and 81.5 (batch 4096) over 8 seeds. It thus reaches a higher success rate with half the gradient steps, no environment interaction, and, on \textit{humanoidmaze-giant}, 25 times less data, which makes it the more sample-efficient of the two.

\subsection{Comparison with TRL}
Transitive RL (TRL) \citep{park2026transitive} reports 79 on \textit{humanoidmaze-giant} in offline goal-conditioned RL, the mean of the evaluations at 800K, 900K, and 1M steps over 4 seeds. It trains for 1M gradient steps with batch 1024, 4$\times$1024 networks, a target network, and $\gamma=0.999$ on the 1B-transition dataset of \citet{park2025horizon}. Its value function is trained with goals relabeled from future states of the same trajectory, so many of its training goals are reached within a few steps. AL-LBLP is trained on the 4M-transition single-task dataset, whose reward is $-1$ at every step until the single evaluation goal is reached, with $4\times512$ networks and no target network. With 1M offline gradient steps it reaches 81.5 (batch 4096, 8 seeds, final evaluation at 1M), which matches TRL with 250 times less data. TRL does not report \textit{antmaze-giant}.

\subsection{Value propagation}
\label{subsec:value_prop}
\begin{figure}[h]
  \centering
  \includegraphics[width=\textwidth]{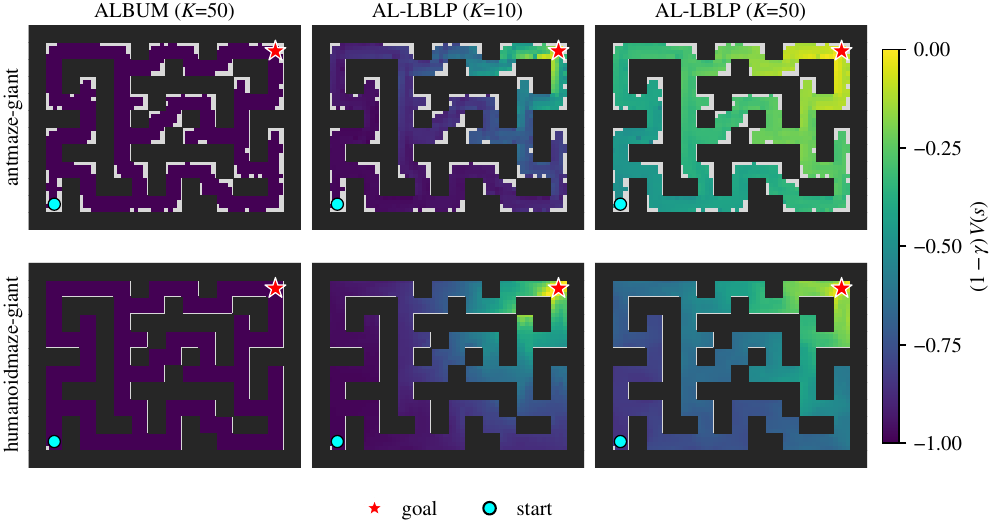}
  \caption{Learned value on task 1 of \textit{antmaze-giant} (top) and \textit{humanoidmaze-giant} (bottom), batch 256: $(1-\gamma)V(s)$ averaged over the dataset states in each $1\times1$ bin (a maze cell is $4\times4$; grey: fewer than 10 states), on a common scale where $-1$ is the floor $-1/(1-\gamma)$. ALBUM and AL-LBLP ($K=50$): $\gamma=0.999$; AL-LBLP ($K=10$): $\gamma=0.995$. Only the upright states are used.}
  \label{fig:value-heatmap}
\end{figure}

\Cref{fig:value-heatmap} shows how far the learned value propagates from the goal. We evaluate the learned $V(s)$ at the states of the dataset (1M on \textit{antmaze-giant}, 4M on \textit{humanoidmaze-giant}), average it over the states whose $xy$ position falls in each $1\times1$ bin of the maze, and plot $(1-\gamma)V(s)$, which is $-1$ at the floor $-1/(1-\gamma)$ of the reward $-1$ per step and $0$ at the goal. Both datasets contain many states in which the agent has fallen. On \textit{antmaze-giant} the ant lies flipped on its back in 21.5\% of the states (its torso's up axis points downward); these states never reach the goal, so their value is correctly at the floor, but they would hide the value of the upright states in the same bin. On \textit{humanoidmaze-giant} the humanoid is down in 39.1\% of the states (root height below 1.0, the healthy-height threshold of the Gymnasium Humanoid task). We therefore sample only the upright states in both environments.

ALBUM does not propagate the value: on both environments $V$ stays at the floor in the whole maze, about $-999$ with a spread below $10^{-3}$, and it almost never succeeds on this task (at most 0.2\%, \cref{tab:longhorizon-tasks}). AL-LBLP propagates the value from the goal through the maze, so that states closer to the goal along the maze have higher values. Raising $K$ to 50 (with $\gamma$ to 0.999) propagates the value farther. With $K=10$ the value saturates far from the goal: on \textit{humanoidmaze-giant}, $(1-\gamma)V$ is already $-0.91$ at the maze cells 20 steps from the goal, while the start is 30 cells away. With $K=50$ the value keeps decreasing toward the start over the whole maze. The $K$-step rows carry this long-range propagation (Theorem~4.5 of \citet{yang2026lifted}), and on these mazes ALBUM violates them by hundreds on the windows that reach the goal, while AL-LBLP keeps them tight (\cref{tab:goal-rows}). We therefore attribute the long-horizon results of \cref{tab:longhorizon} to the $K$-step rows, which AL-LBLP enforces better than ALBUM and which propagate the value farther as $K$ grows.

\newpage
\section{Per-task results}
\label{app:pertask}
\Cref{tab:pertask} lists the five tasks of every environment behind the environment means of \cref{tab:main}.
\begin{table*}[!h]
  \centering
  \caption{Per-task success rate (\%), mean $\pm$ std over 8 seeds (FAC, DROL: mean only); columns and evaluation rules as in \cref{tab:main}. (default): the task used for selection. Bold: within 95\% of the row best.}
  \label{tab:pertask}
    \begin{small}
      \resizebox{\textwidth}{!}{%
      \setlength{\tabcolsep}{3pt}
      \begin{tabular}{lcccccccccc}
        \toprule
        & \multicolumn{4}{c}{Gaussian policy} & \multicolumn{4}{c}{Flow policy} & \multicolumn{2}{c}{Best-of-$N$} \\
        \cmidrule(lr){2-5}\cmidrule(lr){6-9}\cmidrule(lr){10-11}
        Task & IQL & ReBRAC & ALBUM & AL-LBLP & FQL & DROL & FAC & AL-LBLP & IFQL & AL-LBLP \\
        \midrule
        antmaze-large task1 (default) & 48 $\pm$9 & 91 $\pm$10 & 92.5 $\pm$3.1 & 92.5 $\pm$4.9 & 80 $\pm$8 & \textbf{98} & \textbf{94.0} & 89.8 $\pm$5.3 & 24 $\pm$17 & 76.0 $\pm$15.6 \\
        antmaze-large task2 & 42 $\pm$6 & \textbf{88 $\pm$4} & 80.2 $\pm$6.0 & \textbf{85.8 $\pm$3.8} & 57 $\pm$10 & \textbf{84} & \textbf{86.0} & 79.0 $\pm$9.0 & 8 $\pm$3 & 72.2 $\pm$6.5 \\
        antmaze-large task3 & 72 $\pm$7 & 51 $\pm$18 & 91.8 $\pm$2.7 & \textbf{95.8 $\pm$2.7} & \textbf{93 $\pm$3} & 90 & \textbf{97.5} & 90.2 $\pm$7.1 & 52 $\pm$17 & 76.8 $\pm$6.9 \\
        antmaze-large task4 & 51 $\pm$9 & 84 $\pm$7 & \textbf{92.5 $\pm$4.0} & \textbf{93.2 $\pm$4.8} & 80 $\pm$4 & 79 & \textbf{89.5} & \textbf{92.0 $\pm$4.3} & 18 $\pm$8 & 77.2 $\pm$8.6 \\
        antmaze-large task5 & 54 $\pm$22 & 90 $\pm$2 & \textbf{92.0 $\pm$4.7} & \textbf{93.8 $\pm$2.7} & 83 $\pm$4 & \textbf{93} & \textbf{96.0} & \textbf{93.8 $\pm$4.3} & 38 $\pm$18 & 81.8 $\pm$3.6 \\
        \midrule
        antmaze-giant task1 (default) & 0 $\pm$0 & 27 $\pm$22 & 3.5 $\pm$3.6 & \textbf{29.0 $\pm$19.3} & 4 $\pm$5 & \textbf{28} & 6.5 & 8.0 $\pm$5.3 & 0 $\pm$0 & 2.2 $\pm$3.5 \\
        antmaze-giant task2 & 1 $\pm$1 & 16 $\pm$17 & \textbf{76.2 $\pm$7.2} & 62.5 $\pm$10.6 & 9 $\pm$7 & 54 & 37.5 & 46.5 $\pm$8.9 & 0 $\pm$0 & 64.0 $\pm$7.8 \\
        antmaze-giant task3 & 0 $\pm$0 & 34 $\pm$22 & 0.0 $\pm$0.0 & 8.0 $\pm$7.2 & 0 $\pm$1 & \textbf{77} & 0.5 & 1.5 $\pm$1.4 & 0 $\pm$0 & 17.8 $\pm$9.8 \\
        antmaze-giant task4 & 0 $\pm$0 & 5 $\pm$12 & 41.2 $\pm$21.2 & \textbf{64.2 $\pm$9.2} & 14 $\pm$23 & 8 & 20.0 & 38.0 $\pm$8.1 & 0 $\pm$0 & 41.5 $\pm$14.5 \\
        antmaze-giant task5 & 19 $\pm$7 & 49 $\pm$22 & 19.5 $\pm$25.6 & \textbf{74.0 $\pm$18.3} & 16 $\pm$28 & 58 & 50.5 & 53.0 $\pm$12.2 & 13 $\pm$9 & 49.2 $\pm$8.9 \\
        \midrule
        humanoidmaze-medium task1 (default) & 32 $\pm$7 & 16 $\pm$9 & 2.0 $\pm$2.8 & 90.0 $\pm$5.5 & 19 $\pm$12 & 55 & 71.5 & 84.0 $\pm$9.4 & 69 $\pm$19 & \textbf{95.0 $\pm$3.4} \\
        humanoidmaze-medium task2 & 41 $\pm$9 & 18 $\pm$16 & 79.0 $\pm$6.9 & 84.8 $\pm$5.5 & \textbf{94 $\pm$3} & \textbf{95} & 88.0 & 78.8 $\pm$13.1 & 85 $\pm$11 & \textbf{95.8 $\pm$3.5} \\
        humanoidmaze-medium task3 & 25 $\pm$5 & 36 $\pm$13 & 4.8 $\pm$3.7 & \textbf{95.8 $\pm$2.9} & 74 $\pm$18 & 79 & \textbf{95.5} & \textbf{97.8 $\pm$1.7} & 49 $\pm$49 & \textbf{96.2 $\pm$4.2} \\
        humanoidmaze-medium task4 & 0 $\pm$1 & 15 $\pm$16 & 2.2 $\pm$2.5 & 72.5 $\pm$4.6 & 3 $\pm$4 & 17 & 25.0 & 58.0 $\pm$8.8 & 1 $\pm$1 & \textbf{94.5 $\pm$1.8} \\
        humanoidmaze-medium task5 & 66 $\pm$4 & 24 $\pm$20 & 89.5 $\pm$3.1 & 85.0 $\pm$7.9 & \textbf{97 $\pm$2} & \textbf{96} & \textbf{98.0} & 74.2 $\pm$14.0 & \textbf{98 $\pm$2} & \textbf{99.0 $\pm$1.1} \\
        \midrule
        humanoidmaze-large task1 (default) & 3 $\pm$1 & 2 $\pm$1 & 0.0 $\pm$0.0 & 37.0 $\pm$8.5 & 7 $\pm$6 & 20 & 15.0 & 15.2 $\pm$4.8 & 6 $\pm$2 & \textbf{50.0 $\pm$10.6} \\
        humanoidmaze-large task2 & 0 $\pm$0 & 0 $\pm$0 & 0.0 $\pm$0.0 & 1.5 $\pm$2.1 & 0 $\pm$0 & 0 & 0.0 & 0.0 $\pm$0.0 & 0 $\pm$0 & \textbf{3.0 $\pm$4.1} \\
        humanoidmaze-large task3 & 7 $\pm$3 & 8 $\pm$4 & 6.2 $\pm$4.8 & 18.8 $\pm$5.2 & 11 $\pm$7 & 34 & 20.5 & 21.8 $\pm$3.3 & 48 $\pm$10 & \textbf{79.0 $\pm$4.8} \\
        humanoidmaze-large task4 & 1 $\pm$0 & 1 $\pm$1 & 0.5 $\pm$1.3 & 4.8 $\pm$3.0 & 2 $\pm$3 & 1 & 4.5 & 4.8 $\pm$4.5 & 1 $\pm$1 & \textbf{55.0 $\pm$8.2} \\
        humanoidmaze-large task5 & 1 $\pm$1 & 2 $\pm$2 & 1.0 $\pm$1.0 & 25.0 $\pm$6.4 & 1 $\pm$3 & 0 & 1.5 & 20.5 $\pm$10.8 & 0 $\pm$0 & \textbf{51.5 $\pm$3.5} \\
        \midrule
        antsoccer-arena task1 & 14 $\pm$5 & 0 $\pm$0 & 66.0 $\pm$7.0 & 64.2 $\pm$11.4 & 77 $\pm$4 & 76 & \textbf{82.0} & \textbf{83.2 $\pm$6.8} & 61 $\pm$25 & \textbf{85.8 $\pm$4.7} \\
        antsoccer-arena task2 & 17 $\pm$7 & 0 $\pm$1 & 58.0 $\pm$7.8 & 57.5 $\pm$5.6 & 88 $\pm$3 & 87 & \textbf{93.5} & 71.5 $\pm$7.5 & 75 $\pm$3 & 75.2 $\pm$9.4 \\
        antsoccer-arena task3 & 6 $\pm$4 & 0 $\pm$0 & 4.8 $\pm$6.9 & 50.8 $\pm$12.1 & \textbf{61 $\pm$6} & \textbf{60} & \textbf{62.5} & 49.5 $\pm$13.3 & 14 $\pm$22 & 48.8 $\pm$7.7 \\
        antsoccer-arena task4 (default) & 3 $\pm$2 & 0 $\pm$0 & 30.2 $\pm$9.1 & 28.0 $\pm$5.8 & 39 $\pm$6 & 41 & \textbf{53.0} & 45.2 $\pm$10.0 & 16 $\pm$9 & 43.5 $\pm$10.1 \\
        antsoccer-arena task5 & 2 $\pm$2 & 0 $\pm$0 & 21.5 $\pm$5.6 & 41.8 $\pm$10.9 & 36 $\pm$9 & \textbf{60} & 47.5 & 32.8 $\pm$12.3 & 0 $\pm$1 & 27.2 $\pm$7.4 \\
        \midrule
        cube-single task1 & 88 $\pm$3 & 89 $\pm$5 & 84.5 $\pm$10.6 & 88.2 $\pm$5.1 & \textbf{97 $\pm$2} & \textbf{98} & \textbf{99.0} & \textbf{94.2 $\pm$3.6} & 79 $\pm$4 & 92.0 $\pm$2.1 \\
        cube-single task2 (default) & 85 $\pm$8 & 92 $\pm$4 & 89.8 $\pm$6.1 & 93.2 $\pm$4.1 & \textbf{97 $\pm$2} & \textbf{99} & \textbf{100.0} & \textbf{95.2 $\pm$3.5} & 73 $\pm$3 & 94.0 $\pm$3.9 \\
        cube-single task3 & 91 $\pm$5 & 93 $\pm$3 & 90.8 $\pm$4.1 & 94.0 $\pm$3.9 & \textbf{98 $\pm$2} & \textbf{100} & \textbf{100.0} & 94.2 $\pm$3.5 & 88 $\pm$4 & \textbf{95.8 $\pm$2.7} \\
        cube-single task4 & 73 $\pm$6 & 92 $\pm$3 & 84.2 $\pm$6.5 & 90.2 $\pm$4.3 & \textbf{94 $\pm$3} & \textbf{96} & \textbf{98.5} & 86.5 $\pm$3.3 & 79 $\pm$6 & 87.5 $\pm$5.4 \\
        cube-single task5 & 78 $\pm$9 & 87 $\pm$8 & 67.0 $\pm$6.0 & 80.0 $\pm$10.0 & \textbf{93 $\pm$3} & \textbf{96} & \textbf{96.5} & 86.5 $\pm$5.5 & 77 $\pm$7 & 78.2 $\pm$8.1 \\
        \midrule
        cube-double task1 & 27 $\pm$5 & 45 $\pm$6 & 47.2 $\pm$8.8 & 38.2 $\pm$5.9 & 61 $\pm$9 & \textbf{65} & 60.0 & 40.0 $\pm$12.6 & 35 $\pm$9 & 46.0 $\pm$12.2 \\
        cube-double task2 (default) & 1 $\pm$1 & 7 $\pm$3 & 5.2 $\pm$4.7 & 16.8 $\pm$8.9 & \textbf{36 $\pm$6} & 28 & \textbf{37.5} & 19.0 $\pm$5.5 & 9 $\pm$5 & 18.5 $\pm$8.3 \\
        cube-double task3 & 0 $\pm$0 & 4 $\pm$1 & 4.5 $\pm$3.1 & 12.2 $\pm$5.3 & 22 $\pm$5 & 22 & \textbf{31.5} & 11.0 $\pm$4.7 & 8 $\pm$5 & 14.8 $\pm$9.4 \\
        cube-double task4 & 0 $\pm$0 & 1 $\pm$1 & 2.0 $\pm$1.4 & 1.5 $\pm$1.4 & 5 $\pm$2 & \textbf{9} & 4.0 & 1.5 $\pm$1.4 & 1 $\pm$1 & 3.0 $\pm$2.8 \\
        cube-double task5 & 4 $\pm$3 & 4 $\pm$2 & 7.8 $\pm$7.0 & 13.0 $\pm$6.3 & 19 $\pm$10 & 8 & \textbf{32.5} & 16.0 $\pm$6.1 & 17 $\pm$6 & 19.0 $\pm$8.8 \\
        \midrule
        scene task1 & 94 $\pm$3 & \textbf{95 $\pm$2} & 91.8 $\pm$4.9 & \textbf{96.5 $\pm$3.7} & \textbf{100 $\pm$0} & \textbf{100} & \textbf{100.0} & \textbf{99.5 $\pm$0.9} & \textbf{98 $\pm$3} & \textbf{99.8 $\pm$0.7} \\
        scene task2 (default) & 12 $\pm$3 & 50 $\pm$13 & 93.2 $\pm$3.2 & \textbf{97.0 $\pm$1.9} & 76 $\pm$9 & 89 & \textbf{100.0} & \textbf{98.8 $\pm$1.5} & 0 $\pm$0 & \textbf{99.8 $\pm$0.7} \\
        scene task3 & 32 $\pm$7 & 55 $\pm$16 & 70.8 $\pm$11.7 & 86.2 $\pm$9.0 & \textbf{98 $\pm$1} & \textbf{98} & \textbf{97.0} & 89.0 $\pm$5.0 & 54 $\pm$19 & \textbf{96.8 $\pm$3.2} \\
        scene task4 & 0 $\pm$1 & 3 $\pm$3 & \textbf{66.2 $\pm$9.5} & 13.0 $\pm$8.9 & 5 $\pm$1 & 3 & 58.0 & 24.2 $\pm$18.3 & 0 $\pm$0 & 18.0 $\pm$4.0 \\
        scene task5 & 0 $\pm$0 & 0 $\pm$0 & 0.0 $\pm$0.0 & 0.0 $\pm$0.0 & 0 $\pm$0 & 0 & \textbf{1.5} & 0.0 $\pm$0.0 & 0 $\pm$0 & 0.5 $\pm$0.9 \\
        \midrule
        puzzle-3x3 task1 & 33 $\pm$6 & \textbf{97 $\pm$4} & \textbf{96.2 $\pm$4.1} & \textbf{96.8 $\pm$3.5} & 90 $\pm$4 & 93 & \textbf{100.0} & \textbf{100.0 $\pm$0.0} & 94 $\pm$3 & \textbf{99.2 $\pm$1.5} \\
        puzzle-3x3 task2 & 4 $\pm$3 & 1 $\pm$1 & 77.5 $\pm$32.6 & 88.5 $\pm$27.0 & 16 $\pm$5 & 24 & \textbf{100.0} & \textbf{99.0 $\pm$1.9} & 1 $\pm$2 & \textbf{95.2 $\pm$7.5} \\
        puzzle-3x3 task3 & 3 $\pm$2 & 3 $\pm$1 & 39.0 $\pm$16.7 & 74.8 $\pm$15.7 & 10 $\pm$3 & 7 & \textbf{100.0} & \textbf{96.0 $\pm$2.6} & 0 $\pm$0 & 91.5 $\pm$7.1 \\
        puzzle-3x3 task4 (default) & 2 $\pm$1 & 2 $\pm$1 & 93.0 $\pm$7.9 & 86.8 $\pm$11.3 & 16 $\pm$5 & 36 & \textbf{100.0} & 94.2 $\pm$4.9 & 0 $\pm$0 & \textbf{97.8 $\pm$2.7} \\
        puzzle-3x3 task5 & 3 $\pm$2 & 5 $\pm$3 & 0.8 $\pm$1.4 & 82.0 $\pm$12.6 & 16 $\pm$3 & 46 & \textbf{100.0} & \textbf{98.8 $\pm$1.8} & 0 $\pm$0 & \textbf{96.8 $\pm$4.4} \\
        \midrule
        puzzle-4x4 task1 & 12 $\pm$2 & 26 $\pm$4 & 40.2 $\pm$6.4 & 30.8 $\pm$14.0 & 34 $\pm$8 & 39 & \textbf{52.0} & 40.0 $\pm$9.4 & 49 $\pm$9 & 32.8 $\pm$9.2 \\
        puzzle-4x4 task2 & 7 $\pm$4 & 12 $\pm$4 & 11.2 $\pm$3.5 & 4.8 $\pm$4.8 & 16 $\pm$5 & \textbf{20} & 7.5 & 11.2 $\pm$8.2 & 4 $\pm$4 & 12.8 $\pm$4.7 \\
        puzzle-4x4 task3 & 9 $\pm$3 & 15 $\pm$3 & 33.2 $\pm$15.0 & 29.2 $\pm$19.9 & 18 $\pm$5 & 32 & \textbf{62.0} & 42.8 $\pm$17.8 & 50 $\pm$14 & 35.2 $\pm$8.9 \\
        puzzle-4x4 task4 (default) & 5 $\pm$2 & 10 $\pm$3 & 11.8 $\pm$4.7 & 16.2 $\pm$11.1 & 11 $\pm$3 & 16 & \textbf{35.0} & 14.5 $\pm$7.3 & 21 $\pm$11 & 16.5 $\pm$3.8 \\
        puzzle-4x4 task5 & 4 $\pm$1 & 7 $\pm$3 & 6.2 $\pm$3.1 & 5.5 $\pm$4.5 & 7 $\pm$3 & \textbf{17} & 5.0 & 3.2 $\pm$3.0 & 2 $\pm$2 & 7.8 $\pm$5.2 \\
        \bottomrule
      \end{tabular}
      }
    \end{small}
\end{table*}

\newpage
\section{Offline-to-online fine-tuning}
\label{app:off2on}
\Cref{tab:off2on-full} lists the ten default tasks for the three AL-LBLP policies of \cref{fig:combined_results}(a); the online settings are in \cref{app:details}.

\begin{table*}[h]
  \centering
  \caption{Offline-to-online per environment (default task): success (\%) at the offline 1M checkpoint $\to$ after 1M online steps (final evaluation), mean $\pm$ std over 3 seeds. Bold: the best online value.}
  \label{tab:off2on-full}
    \begin{small}
      \begin{tabular}{lccc}
        \toprule
        Environment & Gaussian policy & Flow policy & Best-of-$N$ \\
        \midrule
        antmaze-large & 94.7 $\pm$1.2 $\to$ \textbf{98.7} $\pm$2.3 & 85.3 $\pm$3.1 $\to$ 96.0 $\pm$2.0 & 78.0 $\pm$15.6 $\to$ 97.3 $\pm$3.1 \\
        antmaze-giant & 34.7 $\pm$11.4 $\to$ \textbf{94.7} $\pm$1.2 & 10.0 $\pm$0.0 $\to$ 63.3 $\pm$54.9 & 0.0 $\pm$0.0 $\to$ 93.3 $\pm$2.3 \\
        humanoidmaze-medium & 84.7 $\pm$7.0 $\to$ 99.3 $\pm$1.2 & 90.0 $\pm$6.0 $\to$ \textbf{100.0} $\pm$0.0 & 96.7 $\pm$1.2 $\to$ 98.0 $\pm$2.0 \\
        humanoidmaze-large & 47.3 $\pm$2.3 $\to$ 38.0 $\pm$11.1 & 14.7 $\pm$8.1 $\to$ 23.3 $\pm$9.5 & 59.3 $\pm$7.0 $\to$ \textbf{86.7} $\pm$4.6 \\
        antsoccer-arena & 26.7 $\pm$8.1 $\to$ \textbf{57.3} $\pm$4.2 & 44.0 $\pm$3.5 $\to$ 54.0 $\pm$13.1 & 45.3 $\pm$12.1 $\to$ 38.7 $\pm$6.1 \\
        cube-single & 94.7 $\pm$4.2 $\to$ 90.7 $\pm$6.1 & 94.0 $\pm$6.0 $\to$ 94.0 $\pm$8.7 & 98.0 $\pm$0.0 $\to$ \textbf{97.3} $\pm$2.3 \\
        cube-double & 19.3 $\pm$17.9 $\to$ 29.3 $\pm$5.0 & 15.3 $\pm$4.2 $\to$ 11.3 $\pm$6.4 & 26.7 $\pm$9.2 $\to$ \textbf{42.7} $\pm$27.2 \\
        scene & 96.0 $\pm$3.5 $\to$ 74.7 $\pm$17.9 & 97.3 $\pm$2.3 $\to$ \textbf{93.3} $\pm$2.3 & 98.7 $\pm$1.2 $\to$ 74.0 $\pm$21.6 \\
        puzzle-3x3 & 97.3 $\pm$3.1 $\to$ 33.3 $\pm$57.7 & 96.0 $\pm$2.0 $\to$ \textbf{98.7} $\pm$2.3 & 97.3 $\pm$2.3 $\to$ 98.0 $\pm$3.5 \\
        puzzle-4x4 & 20.0 $\pm$9.2 $\to$ 12.7 $\pm$10.1 & 24.0 $\pm$7.2 $\to$ \textbf{97.3} $\pm$2.3 & 18.7 $\pm$8.3 $\to$ 37.3 $\pm$37.0 \\
        \midrule
        Average & 61.5 $\to$ 62.9 & 57.1 $\to$ 73.1 & 61.9 $\to$ 76.3 \\
        \bottomrule
      \end{tabular}
    \end{small}
\end{table*}

\section{Pixel-based tasks}
\label{app:visual}
\begin{table}[h]
  \centering
  \caption{Pixel-based tasks (\textit{visual-}, task 1): success (\%) at 500k steps, 4 seeds. AL-LBLP (Gaussian policy): final evaluation, mean $\pm$ std, $(\rho,\alpha)$ from $\{0.5,1,2\}\times\{10,30,100\}$ selected on the reported seeds; baselines: reported mean. Bold: within 95\% of the row best.}
  \label{tab:visual}
    \begin{small}
      \setlength{\tabcolsep}{3pt}
      \begin{tabular}{lccccc}
        \toprule
        Environment & IQL & ReBRAC & FQL & FAC & AL-LBLP \\
        \midrule
        cube-single & 70 & \textbf{83} & \textbf{81} & \textbf{82.0} & \textbf{79.5} $\pm$8.2 \\
        cube-double & 34 & 4 & 21 & 48.0 & \textbf{52.5} $\pm$4.1 \\
        scene & \textbf{97} & \textbf{98} & \textbf{98} & \textbf{99.0} & \textbf{95.0} $\pm$2.6 \\
        puzzle-3x3 & 7 & 88 & \textbf{94} & \textbf{95.0} & \textbf{93.0} $\pm$6.2 \\
        puzzle-4x4 & 0 & 26 & 33 & \textbf{56.0} & 3.5 $\pm$4.7 \\
        \midrule
        Average & 41.6 & 59.8 & 65.4 & \textbf{76.0} & 64.7 $\pm$2.6 \\
        \bottomrule
      \end{tabular}
    \end{small}
\end{table}

\begin{figure}[h]
  \centering
  \includegraphics[width=0.7\textwidth]{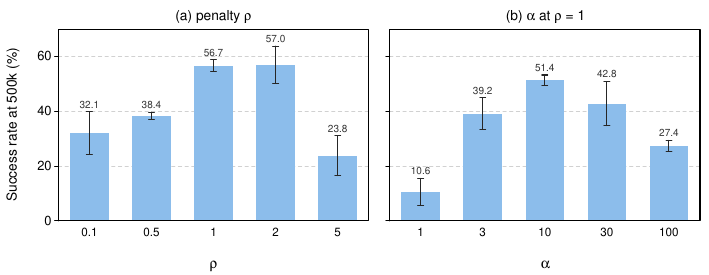}
  \caption{Sensitivity on the five pixel-based manipulation tasks (task 1): five-environment mean success (\%) of the final evaluation at 500k steps, mean $\pm$ std over 4 seeds. (a) Penalty $\rho$, $\alpha\in\{10,30,100\}$ selected per environment. (b) $\alpha$ at $\rho=1$, the same for every environment.}
  \label{fig:visual-sens}
\end{figure}

\Cref{tab:visual} compares AL-LBLP with the Gaussian policy on task 1 of the five pixel-based environments against the values reported for IQL, ReBRAC, FQL, and FAC (settings in \cref{app:details}). AL-LBLP is within 5\% of the best reported value on \textit{visual-scene} (95.0 against 99.0 for FAC), \textit{visual-puzzle-3x3} (93.0 against 95.0 for FAC), and \textit{visual-cube-single} (79.5 against 83 for ReBRAC), is ahead on \textit{visual-cube-double} (52.5 against 48.0 for FAC), and trails on \textit{visual-puzzle-4x4} (3.5 against 56.0 for FAC). Its average, 64.7, is above those of IQL (41.6) and ReBRAC (59.8) and below those of FQL (65.4) and FAC (76.0). Because $(\rho,\alpha)$ is selected on the same four seeds that are reported, the AL-LBLP column is an optimistic estimate. \Cref{fig:visual-sens} shows the five-environment mean as a function of $\rho$ and of $\alpha$.

\newpage
\section{The penalty parameter $\rho$ per environment}
\label{app:rho}
\Cref{tab:rho} lists the environments behind \cref{fig:combined_results}(b).

\begin{table*}[h]
  \centering
  \caption{Effect of $\rho$ per environment (\cref{fig:combined_results}(b)): success (\%) on the default task at 1M, 2M, and 3M steps, mean over 3 seeds. Bold: the best average.}
  \label{tab:rho}
    \begin{small}
      \setlength{\tabcolsep}{4pt}
        \begin{tabular}{lccccccccc}
          \toprule
          & \multicolumn{3}{c}{$\rho=0.1$} & \multicolumn{3}{c}{$\rho=0.5$} & \multicolumn{3}{c}{$\rho=1$} \\
          \cmidrule(lr){2-4}\cmidrule(lr){5-7}\cmidrule(lr){8-10}
          Task & 1M & 2M & 3M & 1M & 2M & 3M & 1M & 2M & 3M \\
          \midrule
          antmaze-large (default) & 95.3 & 92.7 & 91.3 & 95.3 & 95.3 & 97.3 & 87.3 & 92.7 & 94.7 \\
          antmaze-giant (default) & 31.3 & 35.3 & 29.3 & 0.0 & 38.7 & 54.7 & 0.0 & 0.0 & 32.0 \\
          humanoidmaze-med.\ (default) & 91.3 & 90.7 & 90.0 & 82.0 & 92.0 & 97.3 & 48.7 & 76.7 & 94.0 \\
          humanoidmaze-large (default) & 34.7 & 21.3 & 24.7 & 40.0 & 52.0 & 54.0 & 16.7 & 39.3 & 57.3 \\
          antsoccer (default) & 24.0 & 17.3 & 7.3 & 27.3 & 24.0 & 28.0 & 27.3 & 38.0 & 22.0 \\
          cube-single (default) & 93.3 & 88.7 & 90.0 & 94.7 & 92.7 & 92.0 & 93.3 & 96.7 & 94.7 \\
          cube-double (default) & 4.7 & 2.0 & 2.0 & 17.3 & 5.3 & 8.0 & 14.0 & 17.3 & 8.7 \\
          scene (default) & 91.3 & 38.7 & 8.7 & 96.0 & 92.7 & 74.7 & 94.7 & 97.3 & 90.7 \\
          puzzle-3x3 (default) & 94.0 & 82.7 & 92.0 & 72.7 & 75.3 & 80.0 & 80.0 & 86.7 & 77.3 \\
          puzzle-4x4 (default) & 14.7 & 4.7 & 3.3 & 18.0 & 6.7 & 6.0 & 14.0 & 8.7 & 8.7 \\
          \midrule
          Average & \textbf{57.5} & 47.4 & 43.9 & 54.3 & \textbf{57.5} & \textbf{59.2} & 47.6 & 55.3 & 58.0 \\
          \bottomrule
        \end{tabular}
    \end{small}
\end{table*}

\section{Mass identity at 1M steps}
\label{app:mass}
\Cref{tab:mass} evaluates \cref{eq:mass} at 1M steps on the $\rho=0.5$ runs (5 tasks $\times$ 8 seeds; the critic does not depend on $\alpha$), per environment. Its columns are the left-hand side with the effective multipliers $\mu^{\mathrm{eff}}$ (the identity of \cref{prop:mass} as stated), the scalar form $\mu_{\mathrm{B}}\,\mathbb{E}_i[1-\gamma m_i]+2\mu_{\mathrm{K}}$ (the identity with $\mu$ in place of $\mu^{\mathrm{eff}}$, as in \cref{rem:scalar}), the gap $J_{\mathrm K}$ between the two (evaluated over the training split at the 1M checkpoints of the default task, 3 seeds), and the three scalar multipliers at 1M. The target is $\omega_{Q}+\omega_{V}=0.15$. On \textit{antmaze-large}, \textit{humanoidmaze-large}, and \textit{puzzle-4x4}, tasks 4 and 5 have the same goal, so their critics are identical and the 40 runs of these environments contain 8 identical pairs. Both residuals are logged on the training batch every 10k steps. The batch value of the left-hand side has a sampling standard deviation of 0.023 to 0.043 at 256 windows, so single-batch readings are averaged over the last ten logged steps.

\begin{table}[h]
  \centering
  \caption{Mass identity \eqref{eq:mass} at 1M steps (target $\omega_{Q}+\omega_{V}=0.15$), mean $\pm$ std over the 40 runs per environment ($J_{\mathrm K}$: 3 seeds); columns defined in \cref{app:mass}.}
  \label{tab:mass}
    \begin{small}
      \begin{tabular}{lcccccc}
        \toprule
        Environment & LHS ($\mu^{\mathrm{eff}}$) & LHS ($\mu$) & $J_{\mathrm K}$ & $\mu_{\mathrm{B}}$ & $\mu_{\mathrm{K}}$ & $\mu_{\mathrm{E}}$ \\
        \midrule
        antmaze-large & 0.150 $\pm$0.010 & 0.145 $\pm$0.001 & 0.004 $\pm$0.001 & 0.222 $\pm$0.007 & 0.072 $\pm$0.001 & 0.198 $\pm$0.007 \\
        antmaze-giant & 0.150 $\pm$0.009 & 0.144 $\pm$0.001 & 0.010 $\pm$0.006 & 0.252 $\pm$0.016 & 0.071 $\pm$0.001 & 0.229 $\pm$0.016 \\
        humanoidmaze-medium & 0.150 $\pm$0.009 & 0.144 $\pm$0.001 & 0.005 $\pm$0.001 & 0.238 $\pm$0.006 & 0.071 $\pm$0.000 & 0.214 $\pm$0.007 \\
        humanoidmaze-large & 0.148 $\pm$0.008 & 0.145 $\pm$0.001 & 0.004 $\pm$0.000 & 0.232 $\pm$0.015 & 0.072 $\pm$0.001 & 0.208 $\pm$0.015 \\
        antsoccer-arena & 0.151 $\pm$0.011 & 0.139 $\pm$0.001 & 0.009 $\pm$0.000 & 0.350 $\pm$0.036 & 0.068 $\pm$0.001 & 0.328 $\pm$0.037 \\
        cube-single & 0.154 $\pm$0.011 & 0.143 $\pm$0.002 & 0.012 $\pm$0.001 & 0.336 $\pm$0.019 & 0.067 $\pm$0.001 & 0.312 $\pm$0.020 \\
        cube-double & 0.150 $\pm$0.012 & 0.124 $\pm$0.003 & 0.025 $\pm$0.002 & 0.656 $\pm$0.044 & 0.060 $\pm$0.001 & 0.646 $\pm$0.045 \\
        scene & 0.148 $\pm$0.013 & 0.132 $\pm$0.002 & 0.020 $\pm$0.001 & 0.499 $\pm$0.039 & 0.064 $\pm$0.001 & 0.479 $\pm$0.040 \\
        puzzle-3x3 & 0.149 $\pm$0.017 & 0.132 $\pm$0.001 & 0.020 $\pm$0.003 & 0.430 $\pm$0.012 & 0.065 $\pm$0.001 & 0.406 $\pm$0.012 \\
        puzzle-4x4 & 0.149 $\pm$0.014 & 0.115 $\pm$0.005 & 0.037 $\pm$0.004 & 0.721 $\pm$0.058 & 0.056 $\pm$0.003 & 0.708 $\pm$0.060 \\
        \bottomrule
      \end{tabular}
    \end{small}
\end{table}

\paragraph{The identity holds with $\mu^{\mathrm{eff}}$.}
At 1M the left-hand side with $\mu^{\mathrm{eff}}$ is between 0.148 and 0.154 on every environment, a relative residual between $-1.6\%$ and $+2.5\%$ with a seed spread of 5 to 11 points, that is, zero within batch noise. Evaluated on every window of the training split at the 1M checkpoints (3 seeds, default task), it is between 0.133 and 0.158 on every environment; one run (\textit{antmaze-giant}, seed 0) is at 0.083 with large dual pulls, that is, not yet settled. The identity is reached early. At 10k steps the scalar form is still between 0.002 and 0.063, by 100k it is between 0.139 and 0.146 on every environment, and the $\mu^{\mathrm{eff}}$ form stays between 0.133 and 0.163 at every logged checkpoint from 100k on.

\paragraph{The scalar form falls short by $J_{\mathrm K}$.}
The scalar form is below the target on every environment, 0.115 to 0.145 ($-3\%$ to $-23\%$), most on \textit{cube-double} ($-17\%$) and \textit{puzzle-4x4} ($-23\%$). On the checkpoints the difference between the two forms is accounted for mainly by $J_{\mathrm K}$: on \textit{puzzle-4x4} the difference is 0.041 and $J_{\mathrm K}$ is 0.037. The covariance term between $m_i$ and $\mu^{\mathrm{B,eff}}_i$ is at most 0.004 in magnitude, and the remaining dual-pull terms are zero within their seed spread. On the manipulation environments the scalar form drifts down between 100k and 1M (\textit{cube-double} 0.141 to 0.124, \textit{puzzle-4x4} 0.142 to 0.115) while the $\mu^{\mathrm{eff}}$ form stays at the target. This indicates that $J_{\mathrm K}$ grows during training there: in more windows only one of the KQ and KV rows is penalized. We have not examined why.

\begin{figure}[h]
  \centering
  \includegraphics[width=0.9\textwidth]{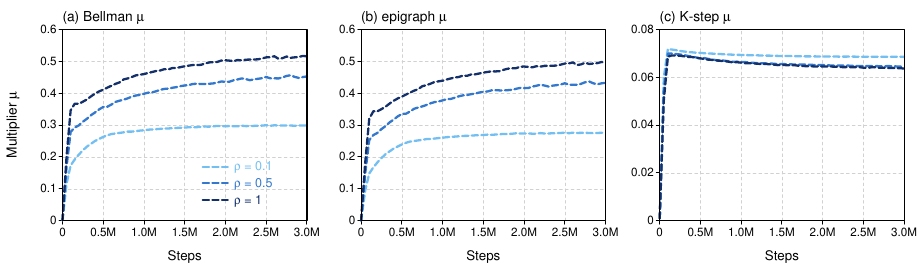}
  \caption{Scalar multipliers $\mu_{\mathrm{B}}$, $\mu_{\mathrm{E}}$, $\mu_{\mathrm{K}}$ over training for $\rho\in\{0.1,0.5,1\}$ (runs of \cref{fig:combined_results}(b)): mean over the 10 default tasks and 3 seeds of the logged value averaged over each 50k steps, starting from the initial value $0.001$.}
  \label{fig:mu-traj}
\end{figure}

\paragraph{Where the multipliers settle.}
$\mathbb{E}_i[1-\gamma m_i]$ is between 0.005 and 0.027, because the success mask is rarely set at a window start, so the one-step row contributes at most 0.006 to the mass at 1M. \Cref{eq:mass} therefore in effect pins the rollout multiplier: $2\mu_{\mathrm{K}}\approx\omega_{Q}+\omega_{V}-J_{\mathrm K}$ up to that contribution. Accordingly $\mu_{\mathrm{K}}$ is between 0.056 and 0.072 on every environment, lowest where $J_{\mathrm K}$ is largest (\textit{puzzle-4x4} 0.056, \textit{cube-double} 0.060). The epigraph multiplier cancels in \cref{eq:mass}. The $b_Q$ equation alone, $\mathbb{E}_i[\mu^{\mathrm{B,eff}}_i]+\mathbb{E}_i[\mu^{\mathrm{KQ,eff}}_i]-\mathbb{E}_i[\mu^{\mathrm{E,eff}}_i]=\omega_{Q}$ (proof of \cref{prop:mass}), fixes the difference $\mu_{\mathrm{B}}-\mu_{\mathrm{E}}\approx\omega_{Q}-\mathbb{E}_i[\mu^{\mathrm{KQ,eff}}_i]$ at the dual fixed point. At 1M the two sides agree to within 0.007 on every environment (for instance 0.025 and 0.025 on \textit{antmaze-large}, 0.010 and 0.007 on \textit{cube-double}), and both bias gradients are zero within batch noise (their means are at most 0.044 in magnitude against $\omega_{Q}=0.1$). The common level of $\mu_{\mathrm{B}}$ and $\mu_{\mathrm{E}}$ is not determined by the two bias equations. It ranges from 0.22 on \textit{antmaze-large} to 0.72 on \textit{puzzle-4x4}, is roughly flat after 100k on the four maze environments, and keeps growing through 1M on \textit{antsoccer-arena} and the manipulation environments (\textit{cube-double} 0.31 at 100k to 0.66 at 1M, \textit{puzzle-4x4} 0.35 to 0.72). We read it as the weight the one-step row needs to hold $Q$ against the epigraph row on each dataset, but we have not tested this reading (\cref{fig:mu-traj}).

\end{document}